\documentclass[final,5p,times,twocolumn]{elsarticle}
\usepackage{caption}
\usepackage{subfigure}
\usepackage[fleqn]{amsmath}
\usepackage{amssymb}
\usepackage{graphicx}
\usepackage{float}
\usepackage{amsthm}
\usepackage{flushend}
\usepackage{lineno}
\usepackage{stfloats}
\usepackage{booktabs}
\usepackage{diagbox}
\usepackage{multirow} 

\journal{International Society for Photogrammetry and Remote Sensing}

\begin{document}

\begin{frontmatter}




\author[inst3]{Xiuwei Zhang}
\cortext[cor1]{Corresponding author. }
\author[inst3]{Yanping Li}
\author[inst3]{Zhaoshuai Qi\corref{cor1}}
\ead{zhaoshuaiqi1206@163.com}

\author[inst4]{Yi Sun}
\author[inst3]{Yanning Zhang}






\affiliation[inst3]{organization={School of Computer Science and Technology},
            addressline={Northwestern Polytechnical University}, 
            city={Xi'an},
            postcode={710072}, 
            state={Shannxi},
            country={China}}

\affiliation[inst4]{organization={School of Cybersecurity},
            addressline={Northwestern Polytechnical University}, 
            city={Xi'an},
            postcode={710072}, 
            state={Shannxi},
            country={China}}

\title{
Learning multi-domain feature relation for visible and Long-wave Infrared image patch matching
}

\begin{abstract}
Recently, learning-based algorithms have achieved promising performance on cross-spectral image patch matching, which, however, is still far from satisfactory for practical application. On the one hand, a lack of large-scale dataset with diverse scenes haunts its further improvement for learning-based algorithms, whose performances and generalization rely heavily on the dataset size and diversity. On the other hand, more emphasis has been put on feature relation in spatial domain whereas the  scale dependency between features has often been ignored, leading to performance degeneration especially when encountering significant appearance variations for cross-spectral patches. To address these issues, we publish, to be best of our knowledge, the largest visible and Long-wave Infrared (LWIR) image patch matching dataset, termed VL-CMIM, which contains 1300 pairs of strictly aligned visible and LWIR images and over 2 million patch pairs covering diverse scenes such as asteroid, field, country, build, street and water. In addition, a multi-domain feature relation learning network (MD-FRN) is proposed. Input by the features extracted from a four-branch network, both feature relations in spatial and  scale domains are learned via a spatial correlation module (SCM) and multi-scale adaptive  aggregation module (MSAG), respectively. To further aggregate the multi-domain relations, a deep domain interactive mechanism (DIM) is applied, where the learnt spatial-relation and scale-relation features are exchanged and further input into MSCRM and SCM. This mechanism allows our model to learn interactive cross-domain feature relations, leading to improved robustness to significant appearance changes due to different modality. Evaluation on the VL-CMIM and other public cross-spectral datasets demonstrates the superior performance of our model on both matching accuracy and generalization against the state of the art. Especially, the FPR95 has been improved by a large margin from 7.35\% to 0.78\% for the Optical-SAR dataset.
\end{abstract}



\begin{keyword}

Multi-modal Image Registration \sep Deep Interactive Integration \sep Visible and Long-wave Infrared Image Registration Dataset
\end{keyword}

\end{frontmatter}


\section{Introduction}
\label{Introduction}
Image patch matching is a fundamental task in computer vision and is widely used in many applications such as image registration \cite{lee2021cnn, fan2016new}, image reconstruction \cite{parameswaran2016patch, wang2017exemplar}, image retrieval  \cite{wang2020deep, setumin2020canonical} and other fields. 
By comparing and matching local patches in images, it is possible to identify and establish correspondences between different image regions. In contrast to single-spectral image patch matching, which may encounter differences from changes of the viewpoint, illumination, the cross-spectral image patching is much more challenging which need to account for additional and more complicated appearance variation from different spectral images. 

Traditional algorithms mainly rely on handcrafted features extracted from distinct image in the pair, from SIFT \cite{lowe2004distinctive}, PCA-SIFT \cite{ke2004pca}, SURF \cite{bay2006surf}, and SSIF \cite{liu2008simplified} to multi-modal image matching  \cite{chen2009real, hossain2011improved, aguilera2012multispectral}, of which the Euclidean or Cosine distance are evaluated for similarity assessment, determining whether images in the pair are matched to each other. While scale and rotation-induced changes have been accounted for in some degree, these low-level local features and their associating descriptors still suffer in significant appearance changes, especially for the case of cross-spectral patches.

\begin{figure*}[ht]
  \centering
  \begin{minipage}[b]{0.45\textwidth}
    \centering
    \subfigure[]{\includegraphics[width=3.4in]{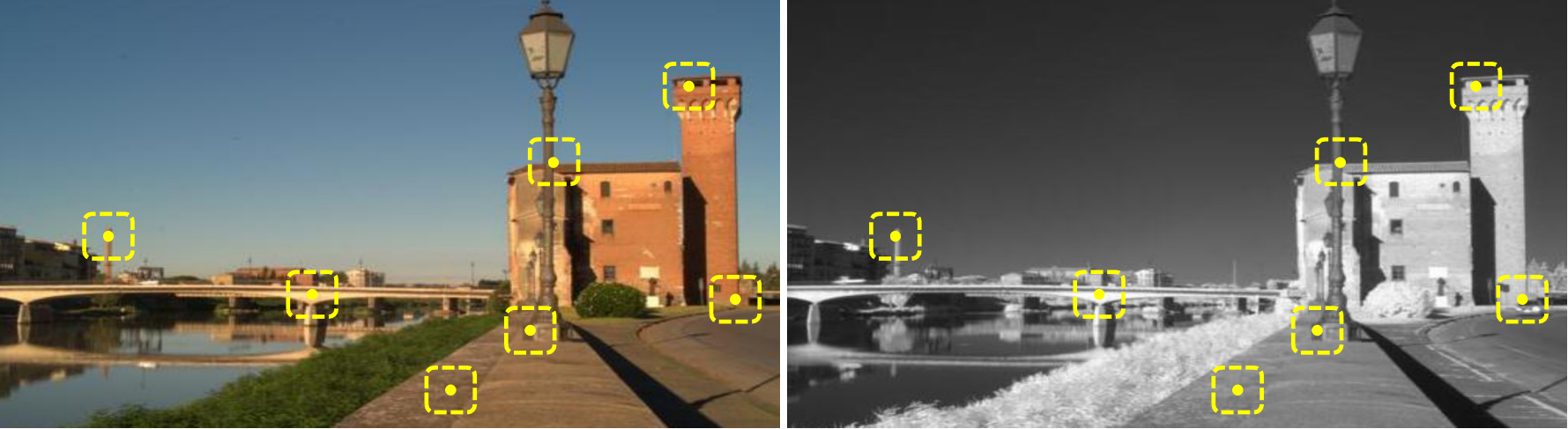}}
    
    \vspace{0 cm} 
    
    \subfigure[]{\includegraphics[width=3.4in]{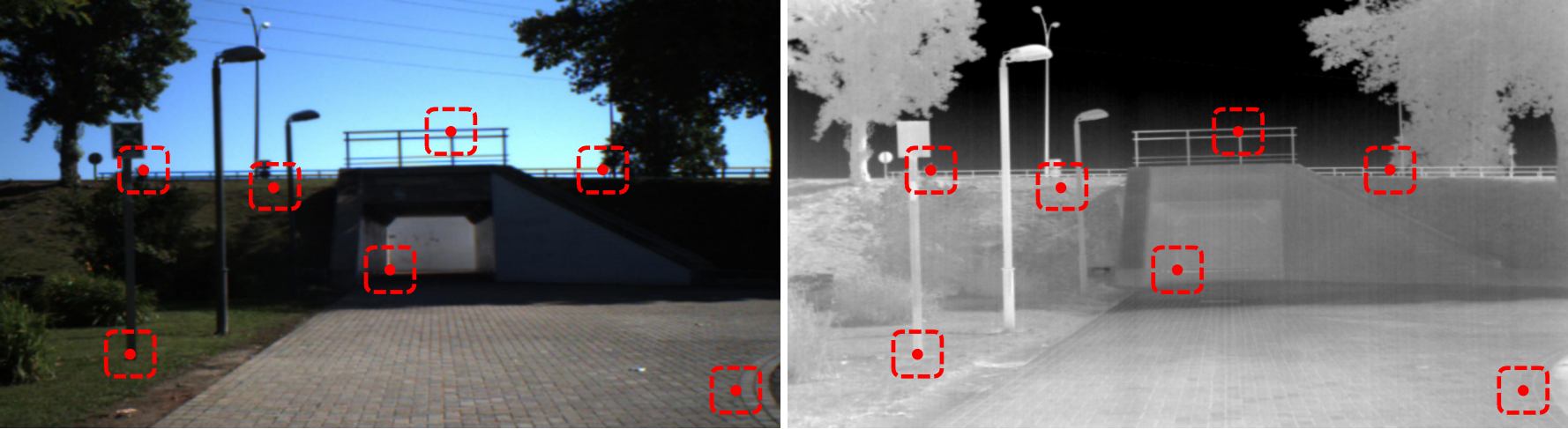}}
  \end{minipage}
  \hfill
  \begin{minipage}[b]{0.52\textwidth}
    \centering
    \subfigure[]{\includegraphics[width=3.7in]{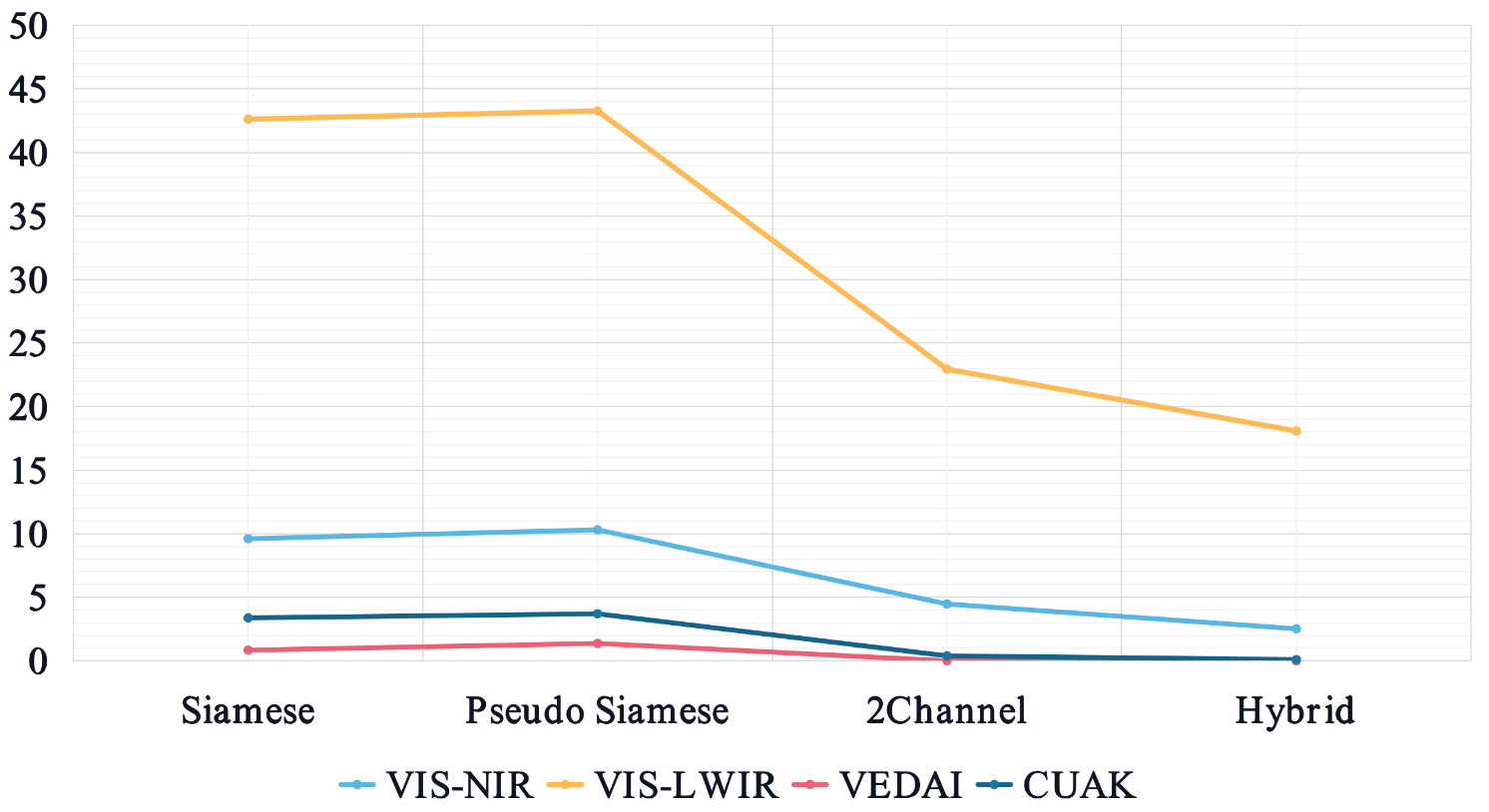}}
  \end{minipage}
  
  \caption{
  (a) shows the Visible image (VIS) and its corresponding near-infrared image (NIR) from the RGB-NIR dataset \cite{brown2011multi}.  (b) shows the Visible image (VIS) alongside its corresponding long-wave infrared image (LWIR) from the LWIR-RGB dataset \cite{aguilera2015lghd}. The matched optical (left) and LWIR (right) images differ by significant appearance changes due to the dissimilar physical characteristics captured by the different sensors. The wavelength range for these images is as follows: VIS: 0.4-0.7 $\mu$m, NIR: 0.75-1.4 $\mu$m, and LWIR: 8-15 $\mu$m. The spectral band between VIS-LWIR is considerably wider than that between VIS-NIR, leading to more pronounced appearance differences between them. (c) demonstrates the average FPR95 values (where lower values indicate better performance) of four methods: Siamese \cite{simo2015discriminative}, Pseudo-Siamese \cite{zagoruyko2015learning}, 2Channel \cite{zagoruyko2015learning}, and Hybrid \cite{baruch2021joint}, across four datasets: VIS-NIR \cite{brown2011multi}, VeDAI \cite{razakarivony2016vehicle}, CUHK \cite{wang2008face} and VIS-LWIR \cite{aguilera2015lghd}. It is observed that these methods perform reasonably well on the VIS-NIR, VeDAI and CUAK datasets. However, when applied to the VIS-LWIR dataset, the values remain relatively high.}
  \label{int1}
\end{figure*}

\begin{table*}[ht]
    \renewcommand\arraystretch{2}
    \centering
    \caption{Summary of Visible and Infrared  Image Pair Datasets for Existing Image(Patch) Matching Tasks.}
    \resizebox{\textwidth}{!}{
    \begin{tabular}{c | c | c | c | c | c}
    \hline
      & Number  & Source & Resolution & Aligned & Observation target and scene description \\
    \hline
      VIS-LWIR \cite{2015LGHD} & 44 & Autonomous University of Barcelona & 639 $\times$ 431 & \checkmark & The campus of the Autonomous University of Barcelona \\
    \hline
      OSU \cite{2007Background} &	6 &	James W. Davis	& 320 $\times$ 240	& \checkmark & Pedestrian intersection on the Ohio State University campus\\
    \hline
      AIC\cite{OCTEC}	& 1 & University College Dublin & 320 $\times$ 240  &  \checkmark &	Outdoor static background at night\\
    \hline
      ITIV\cite{imagefusion} & 3 & Torabi A	& 320 $\times$ 240 &	\checkmark & Indoor scenes, pedestrian targets\\
    \hline
      OCTEC\cite{OCTEC} &	1 &	David Dwyer of OCTEC & 640 $\times$ 480 & \checkmark & Buildings and people covered by smoke grenade with static background outdoors during daytime\\
    \hline
      TNO \cite{2014TNO} & 5 & Alexander Toet & 505 $\times$ 510 &	\checkmark & Contains multispectral nighttime imagery of different military scenes, registered with different multi-band camera systems.\\
    \hline
      IRIS\cite{IRIS} & 31 & Besma Abidi of IRIS Labs & 320 $\times$ 240 &	&	Indoor environment 31 face images of people in different poses, lighting and expressions\\
    \hline
      RoadScene \cite{xu2020fusiondn} & 221 & Electronic Information School, Wuhan University & 500 $\times$ 329 & \checkmark & Containing rich scenes such as roads, vehicles, pedestrians and so on\\
    \hline
     \textbf{VL-CMIM} & 1300 & School of Computer Science, Northwestern Polytechnical University & 1920 $\times$ 1080 & \checkmark & Consists of 2600 images in 6 categories: Asteroid, build, country, field, street and water\\
    \hline
    \end{tabular}}
    \label{ref1}
\end{table*}
 
 In contrast to the traditional algorithm, the learning-based algorithms have demonstrated a powerful capacity for high-level feature extraction and similarity measurement, achieving impressive performance even on cross-spectral patch matching. They can generally fall into two groups: descriptor learning \cite{simo2015discriminative,balntas2016pn,savinov2017quad,tian2017l2,mishchuk2017working,quan2019afd} and metric learning \cite{zagoruyko2015learning,han2015matchnet,kumar2016learning,baruch2021joint,moreshet2021paying}.  
 Despite the success of existing methods like the hybrid approach \cite{baruch2021joint} in achieving good results on specific datasets (Fig. \ref{int1}), they still face challenges in matching visible-light to thermal infrared images. 
 One of the main reasons for this limitation is the scarcity of large-scale cross-spectral patch matching datasets, especially for visible and long-wave infrared (LWIR) image patch pairs, significantly hinders the ability of data-driven algorithms to generalize and enhance accuracy effectively. 
 As shown in Table \ref{ref1}, existing datasets, such as RoadScene \cite{xu2020fusiondn}, IRIS \cite{IRIS} and VIS-LWIR\cite{aguilera2015lghd}, only account for small-scale (a maximum of 221 pairs), low-resolution image pairs from simple scenes such as the campus of university, roads, vehicles and pedestrians and single-view (mostly from the ground view). The limited size and diversity of scenes and views of these datasets hinges the further improvement of learning-based algorithm on generalization and matching accuracy. Secondly, while high-level \cite{yu2023efficient,quan2021multi} or multi-level\cite{yu2022multi,quan2019afd} features have strong invariance for appearance changes, but they only focus on learning invariant and discriminative features for individual image patches that are based on image content.
 However, when it comes to the matching task, it is essential to predict the relationship between two image patches, determining whether they are similar (matching) or dissimilar (non-matching). Therefore, individual feature learning methods based solely on image content are not suitable for addressing the matching problem effectively.
 More specifically, instead of elaborate learning more representative features, AFD-Net \cite{quan2019afd} and MFD-Net \cite{yu2022multi} try to learning discriminative feature relations by aggregating multi-level or multi-branch feature differences, and achieved promising performance. 
 To further consider consistent feature relation between features, EFR-Net \cite{yu2023efficient} and MRAN \cite{quan2021multi} extend the single-relation learning algorithms including AFD-Net \cite{quan2019afd} and MFD-Net \cite{yu2022multi}, and learn a (attentioned) fused relation of concatenation, product and difference, to further improve the performance. However, only features relations in the spatial domain are considered in these works.
 We observe that there is also strong channel dependency between features. This feature relation in channel domain provide additional cues and cross-spectral invariance for the matching, which, however, is often ignored in previous works, leaving rooms for improvement. 

To address the above issues, we construct, to our knowledge, the largest visible and LWIR image patch matching dataset to date, termed VL-CMIM, which, we hope, can serve as a useful research benchmark for cross-spectral image patch matching, especially for ones with significant appearance variation in visible and LWIR patch pairs. In addition, a multi-domain feature relation learning network (MD-FRN), where not only feature relation in spatial domain but also that in channel domain are learnt, leading to improved matching accuracy. Specifically, build upon a four-branch feature extraction network (FB-FEN), two parallel modules, the spatial correlation module (SCM) and multi-scale channel relation learning module (MSCRM), are constructed. SCM is responsible for learning spatial-domain feature relation by correlating features extracted from FB-FEN, and MSCRM tries to learn channel-domain dependency between features. These spatial-channel relation features are then fused via a deep domain interactive mechanism (DIM). By input into MSCRM and SCM iteratively, this mechanism allows our model to learn interactive cross-domain feature relations, leading to improved performance on accuracy and generalization. The main contribution of this work can be summarized as below.

1) The visible and LWIR image patch matching dataset VL-CMIM is proposed. To the best of our knowledge, it is the largest dataset to date in this very specific field. It contains 1300 pairs of high-resolution (1920x1080) visible and LWIR images, over 2 million patch pairs covering diverse scenes such as asteroid, field, country, build, street and water. Moreover, both ground and aerial views are also covered, further improving the diversity of the dataset. VL-CMIM can be served as a useful research benchmark for cross-spectral image patch matching, and the improved size and diversity of dataset will facilitate the development and generalization of learning-based algorithms.

2) Rather than only learning feature relation in spatial domain, MD-FRN is proposed. In contrast to previous works, which only learn feature relations in spatial domain, MD-FRN tries to learns relations in both spatial and channel domain. Extensive results on VL-CMIM and other existing cross-spectral datasets have demonstrated the superiority of our model, where the FPR95 has been improved significantly by a large margin from 7.35\% to 0.78\% for the Optical-SAR dataset. 

The remaining of the paper is arranged as follows. Related works are discussed in Sec. \ref{relate}. The details of the proposed dataset and algorithm are presented in Secs. \ref{vl-cmim} and \ref{scca}, respectively, followed by comparison results of state of the art and our algorithm in Sec. \ref{exper}. The conclusion is drawn in Sec. \ref{conclusion}. 

\begin{figure*}[ht]
    \centering
    \subfigure[Binocular camera]{
        \includegraphics[width=1in]{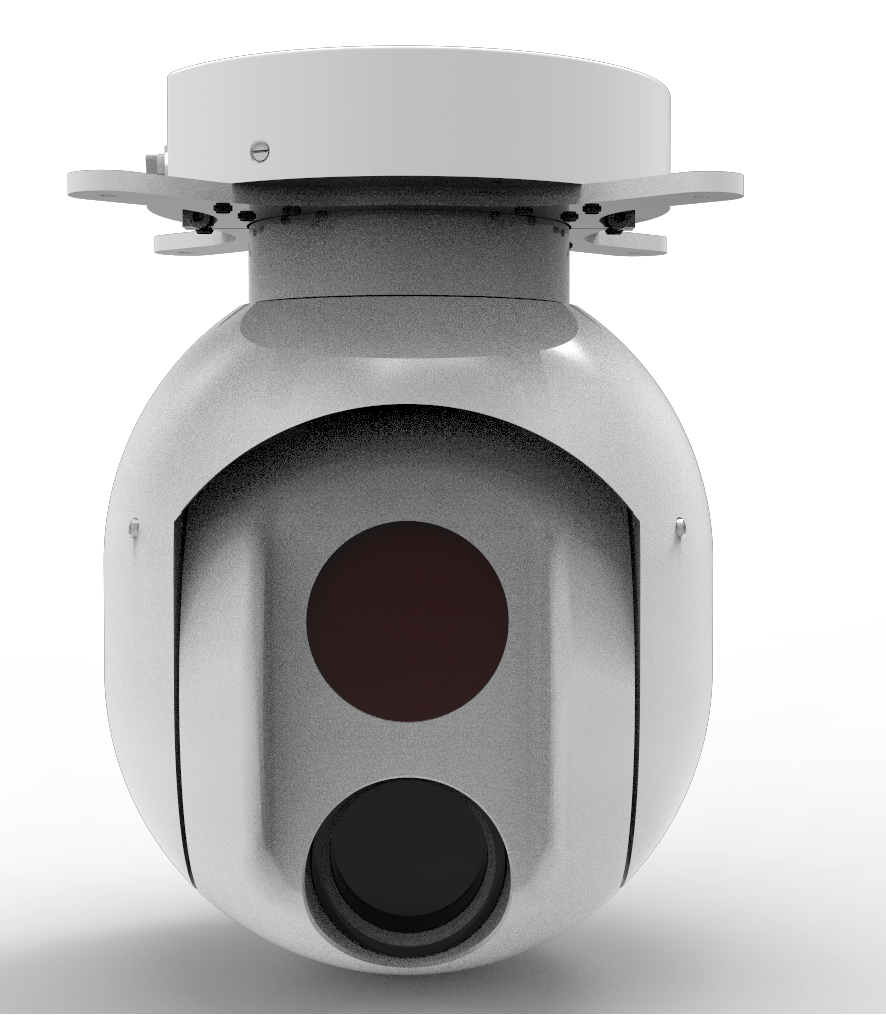}
        \label{img1}
    }
    \subfigure[Image captured in visible light ]{
        \includegraphics[width=2in]{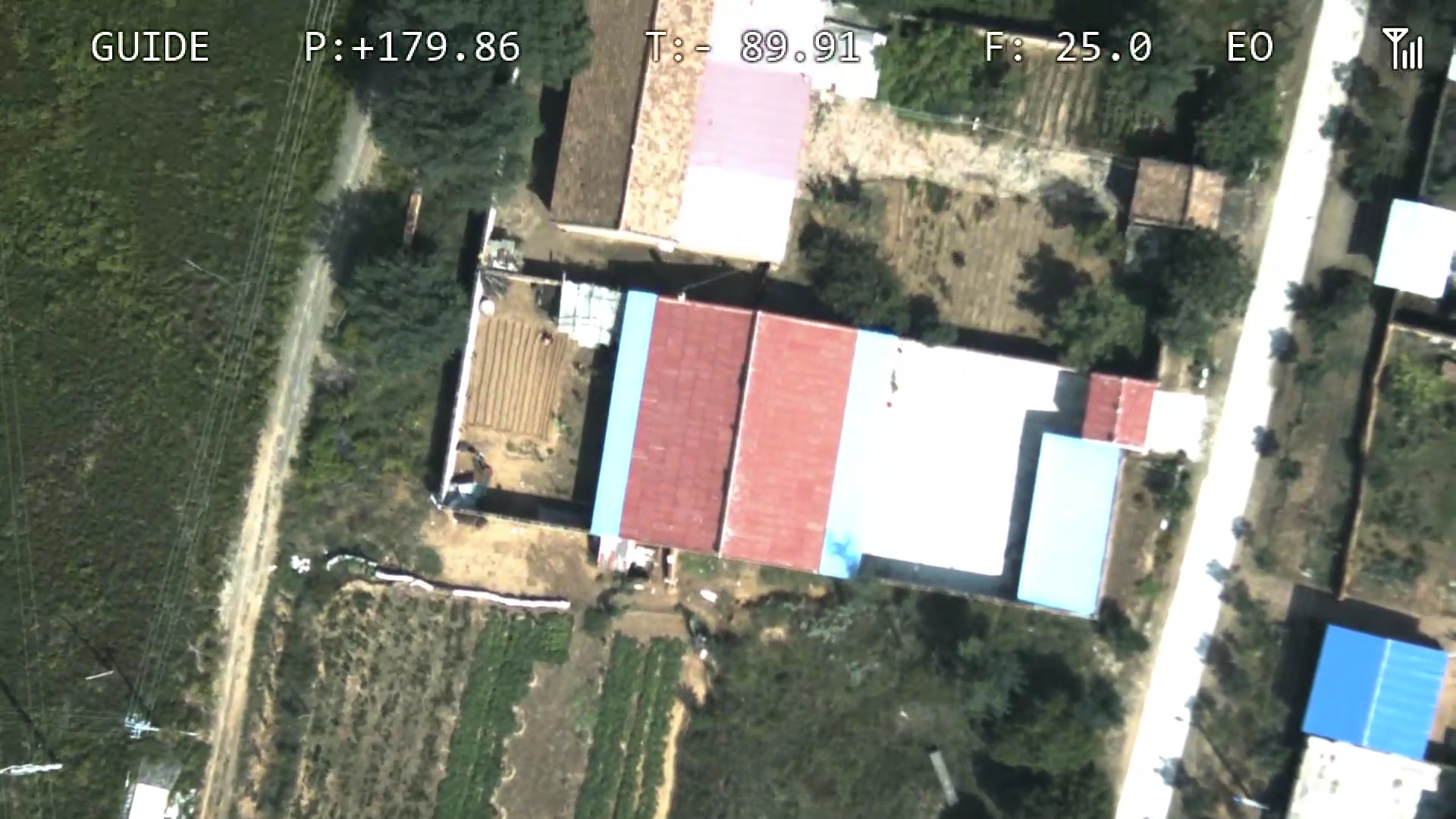}
        \label{img1}
    }
    \subfigure[Image captured in Infrared]{
    	\includegraphics[width=1.4in]{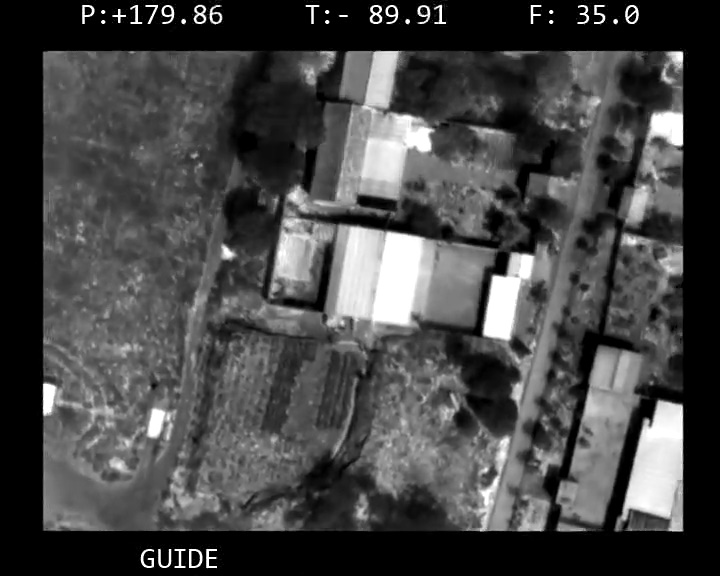}
        \label{img3}
    }
    \caption{The camera used to collect experimental data and the resulting visible light and infrared images, where (a) shows an FAIS-140-04 visible light-thermal infrared binocular camera. Images captured in visible and infrared light are shown in (b) and (c), respectively. }
    \label{cam}
\end{figure*}

\section{Related work} \label{relate}

Image matching algorithms have been a popular research topic for a long time, and the advancements in deep learning have recently renewed interest in the field. Deep learning has significantly improved the performance of image matching algorithms by providing robust image features extracted through Convolutional Neural Networks (CNNs). Consequently, CNN-based algorithms have achieved state-of-the-art results in multimodal image matching tasks.

In general, existing image matching methods can be categorized into two main approaches: metric learning and descriptor learning.

\textbf{Descriptor Learning.}  The goal of descriptor learning is to learn a representation that can enable the two matched features as close as possible, while non-matched features are far apart. Descriptor learning is usually performed using cropped local patches centered on the detected keypoints. It is also known as patch matching. 

Siamese network \cite{firmenichy2011multispectral}, a precursor to descriptor learning, used two branches of convolutional neural network with the same structure and shared weights to learn discriminative features for comparing pairs of image patches. It is optimized by hinge embedding loss, in which the distance between matching patches is expected to approach zero and the distance between non-matching patches is as large as possible or greater than the preselected distance threshold. Unlike the pairwise comparison, PN-Net \cite{balntas2016pn} trained the network with positive and negative pairs consisting of triple patches. It is optimized by a new loss function SoftPN, which has faster convergence and lower error compared to Hinge loss and SoftMax. In contrast to the strict constraint on absolute distance in pairwise networks, triplet comparison emphasizes relative distance, which only needs that the feature distance of matching pairs be smaller than the feature distance of nonmatching pairs. Aguilera et al. \cite{savinov2017quad} proposed Quadruplet Network, which directly applied PN-Net to the cross-spectral VIS-NIR image patch matching problem. 

However, due to the quadratic or cubic sample size, it is extremely difficult to use all samples for optimization, and considering that a random sampling method will introduce many simple samples with ineffective information for optimization, an effective sampling method is required. L2-Net \cite{tian2017l2} applied a progressive sampling strategy to optimize the relative distance-based loss function in the Euclidean space. HardNet \cite{mishchuk2017working} achieves better improvement than L2Net by using a simple hinge triplet loss with the hardest sample mining strategy. To overcone the  overfitting problem in siamese network and triplet network, Vijay Kumar et al. \cite{kumar2016learning} proposed a global loss function that can be applied to mini-batches, which improves the  generalization capability of the model by minimizing the overall classification error in the training set. Quan ei al. \cite{quan2019afd} proposed the AFD-Net that aggregated feature difference learning for cross spectral image patch matching. 

\begin{figure*}[ht]
    \centering
    \subfigure[Space-time alignment registration image]{
        \includegraphics[width=2.8in]{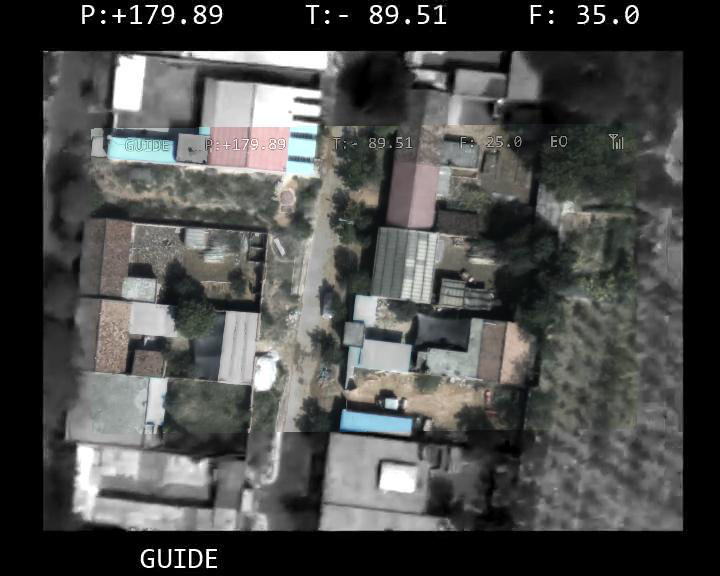}
        \label{img1}
    }
    \subfigure[Checkerboard registration image]{
	      \includegraphics[width=3in]{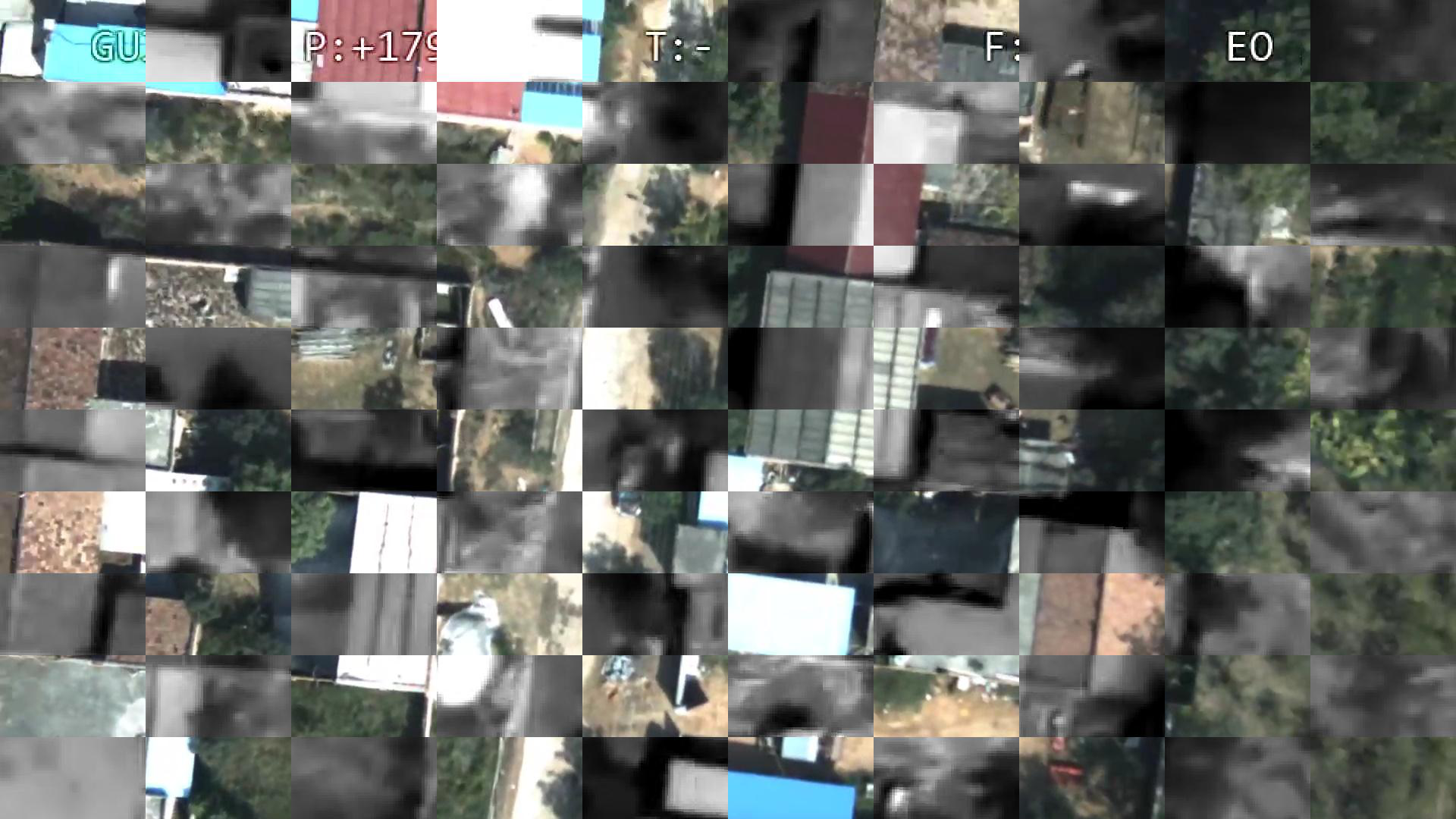}
        \label{img2}
    }
    \quad   
    \subfigure[Visible light image after registration]{
    	\includegraphics[width=3in]{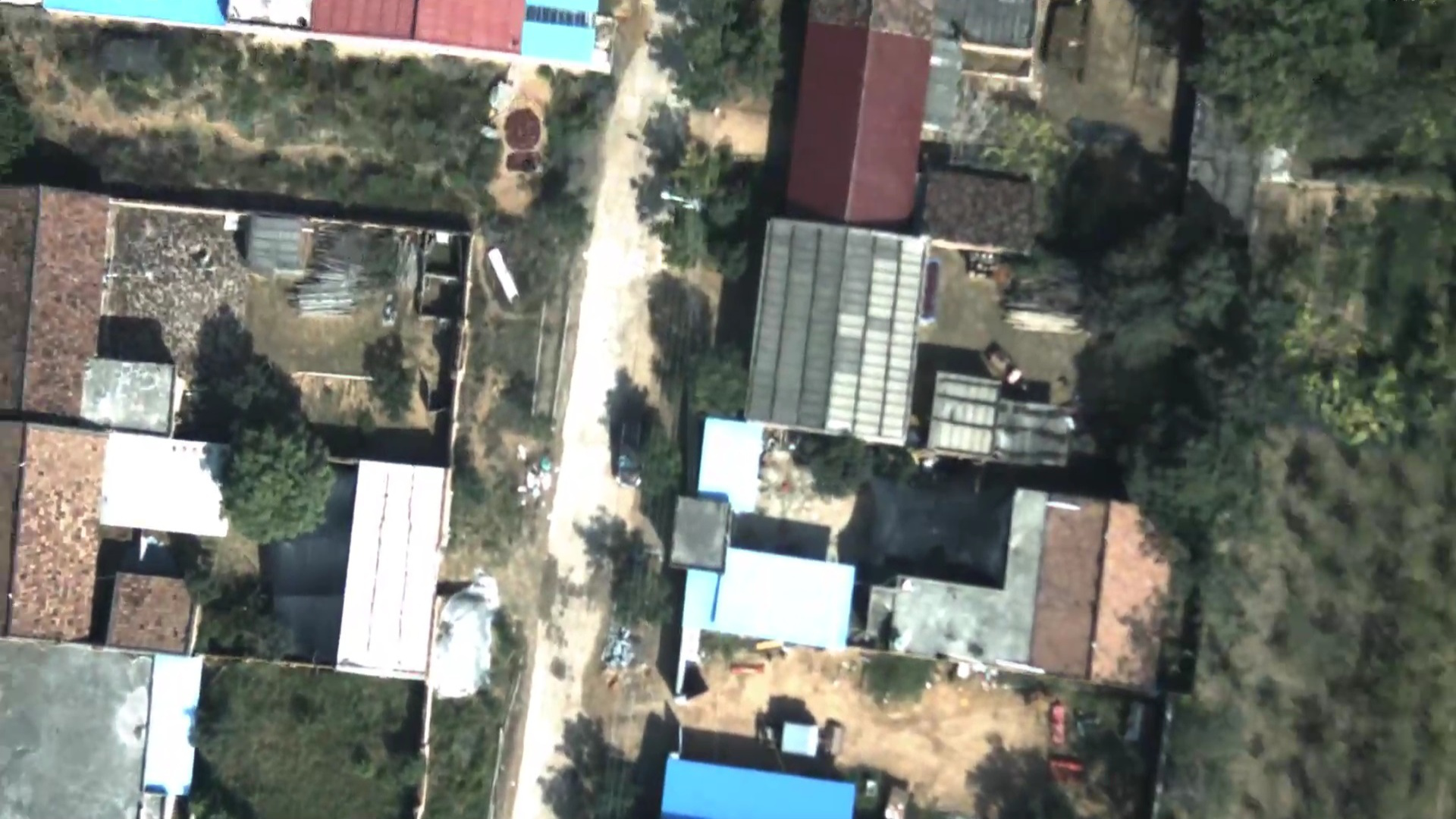}
        \label{img3}
    }
    \subfigure[Infrared image after registration]{
	      \includegraphics[width=3in]{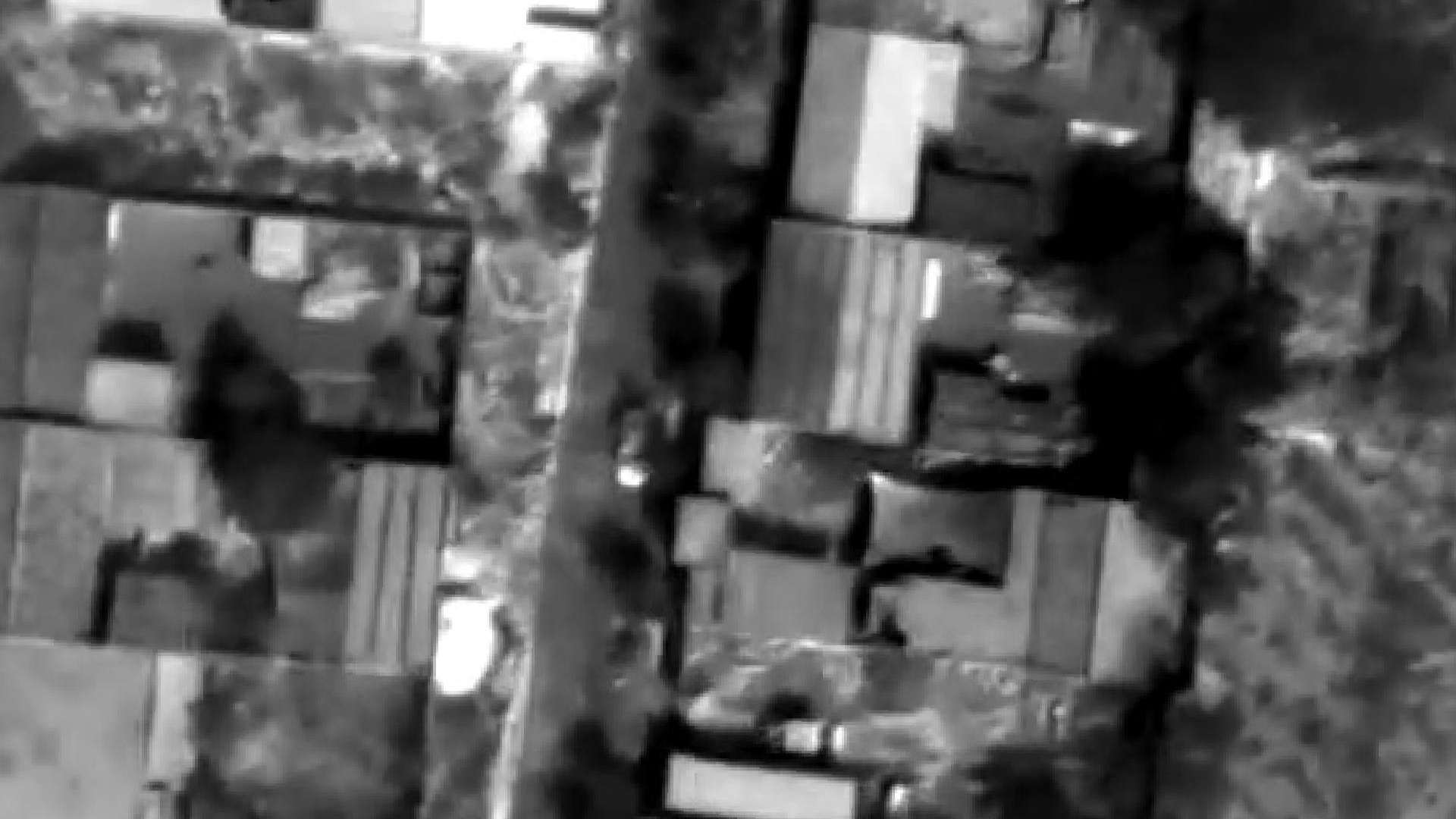}
        \label{img4}
    }
    \caption{The registration results of visible light and thermal infrared images aligned in time and space.}
    \label{reg}
\end{figure*} 

\textbf{Metric Learning.}  Metric learning methods use raw patches or the generated feature descriptors as input to learn discriminative metrics for similarity measurement. It converts the matching task into a binary classification task, and outputs the matching label of the image pairs. 

MatchNet \cite{han2015matchnet} extracts features from image patches through deep convolutional network and uses three fully convolutional layers to calculate the similarity between the extracted features. Under the cross-entropy loss, the image patch matching problem is transformed into a classification problem. Zagoruyko et al. \cite{zagoruyko2015learning} proposed DeepCompare for accessing the similarity between the pairs of image patches using CNNs. They compared different neural network architectures, including Siamese, pseudo-Siamese, and 2-Channel networks. Additionally, they enhanced the baseline models by incorporating the central-surround two-stream network and spatial pyramid pooling (SPP) network to improve performance.  Similar to the 2-channel network that merges two image patches in pixel level, Quan et al. \cite{quan2019cross} proposed the SCFDM that splices two image patches along the spatial dimension. Hybrid \cite{baruch2021joint} combines Siamese and pre-Siamese build four-branch network for multi-modal image matching. Moreshet et al. \cite{moreshet2021paying} used multiscale siamese network extract feature map, combined with transformer to obtain image global information and improve network performance. 

It should be noted that metric learning methods can evaluate descriptor similarity, making them suitable for both unimodal and multimodal image matching tasks. However, multimodal images exhibit complex feature relationships, which pose challenges for metric learning networks in adapting to these variations. Existing multimodal image matching networks, such as SCFDM \cite{quan2019cross}, Hybrid \cite{baruch2021joint}, and Moreshet \cite{moreshet2021paying}, have not fully explored the deep interactions and correlations among multimodal features. To address this limitation, we propose an innovative cross-modal fusion mechanism that effectively integrates feature representations from different modalities by extensively analyzing the feature interaction relationships between modalities. This fusion mechanism captures richer and more accurate feature information, resulting in a significant enhancement of multimodal image matching performance.

\section{VIS-LWIR Cross-Modal Image Patch Matching (VL-CMIM) Dataset} \label{vl-cmim}

\subsection{Dataset Construction}

\textbf{Image capture.} 
There are two methods available for capturing: using a binocular camera or employing simulation software.

The camera equipment we use is FAIS-140-04, a binocular camera platform that consists of a visible light camera and a long-wave infrared camera. We capture images from different locations, such as: Northwestern Polytechnical University campus, Dali country, Bailuyuan Village and Yellow River. Some details about the camera are shown in Fig. \ref{cam}. Another type of simulation software is designed to simulate the process of a vehicle navigating through space to approach and orbit an asteroid while recording image data in video format in both visible and infrared light forms.

\begin{figure*}[ht]
    \centering
    \noindent

    {\includegraphics[width=1in]{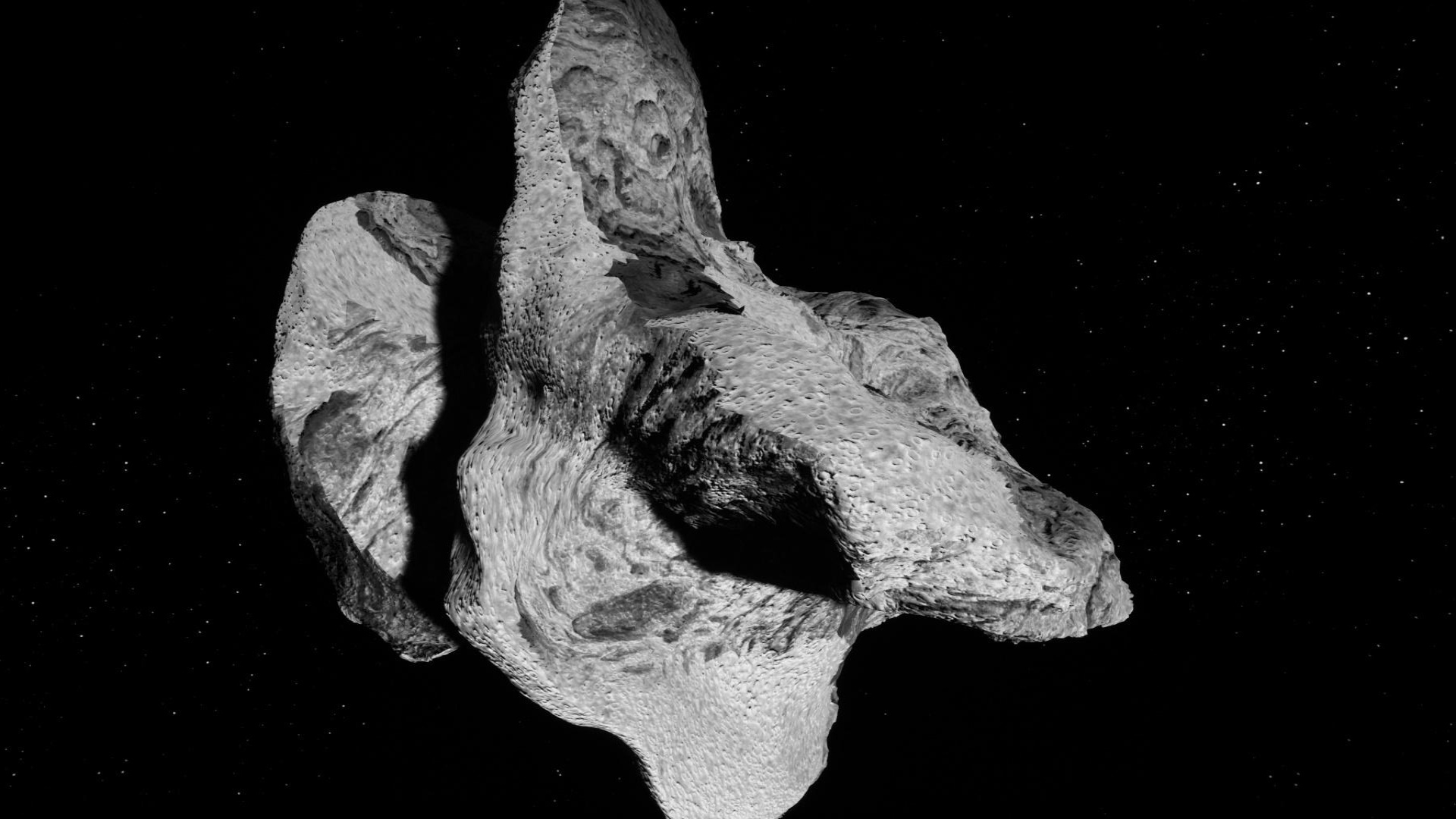}}
    {\includegraphics[width=1in]{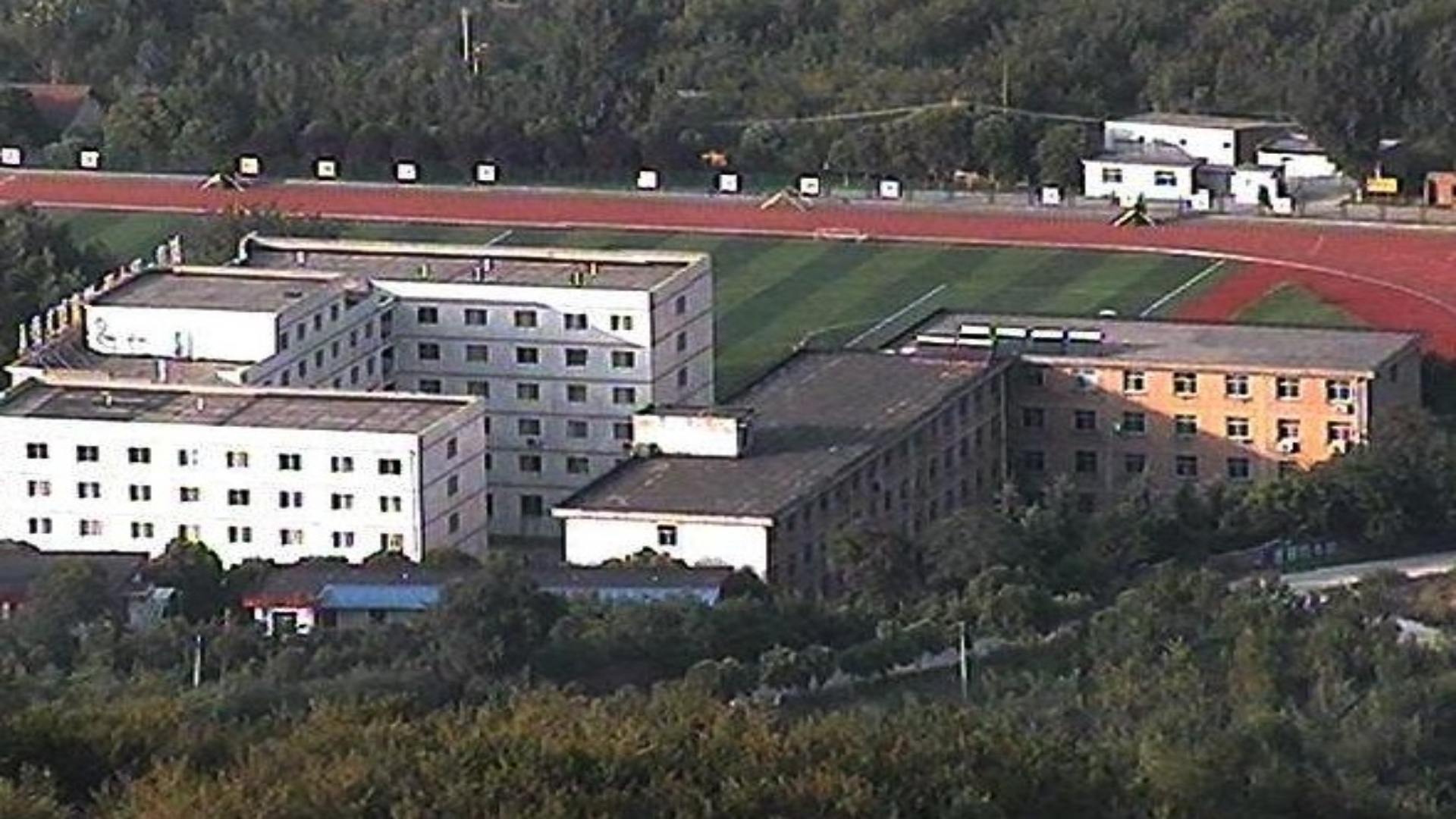}}
    {\includegraphics[width=1in]{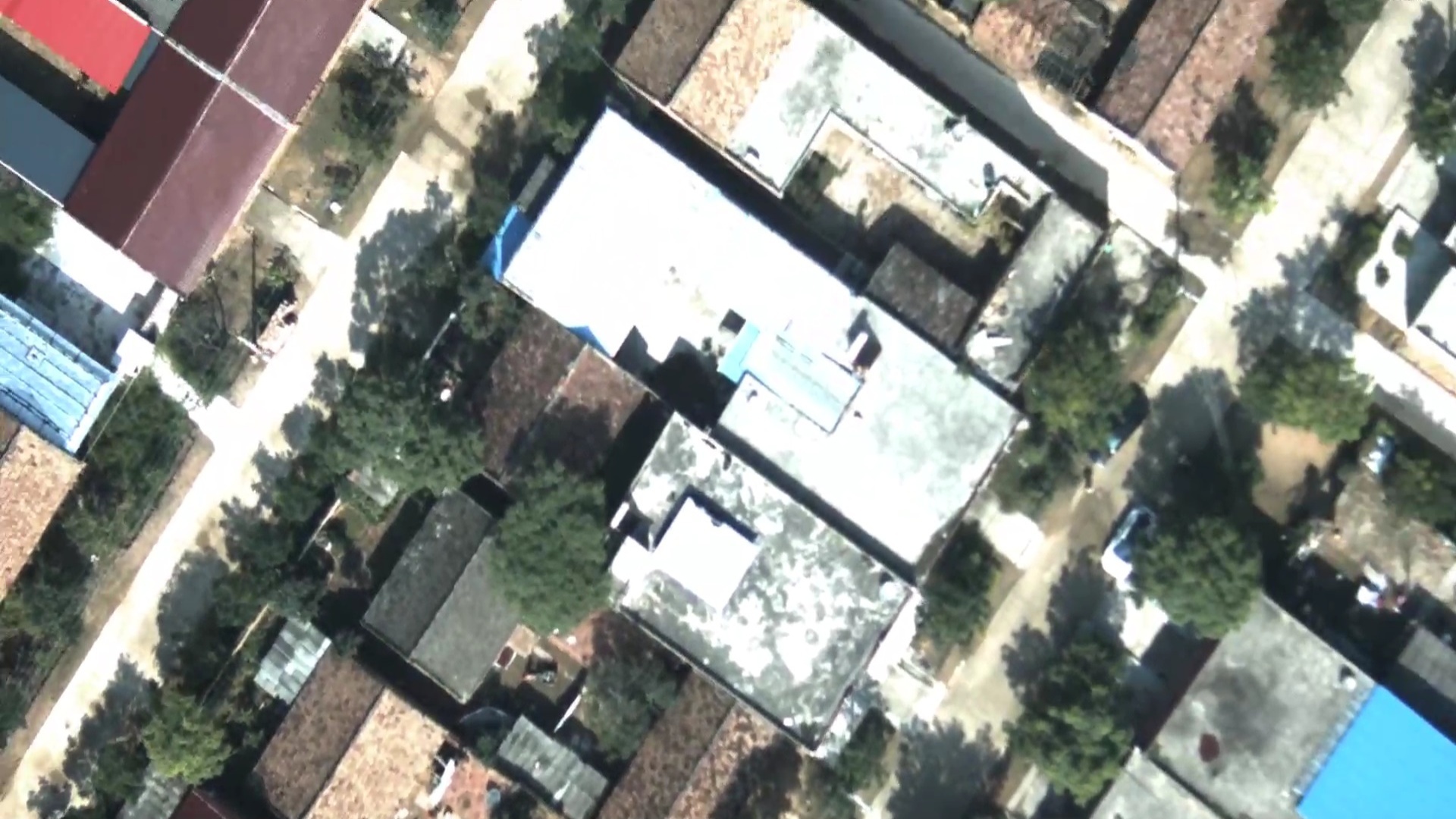}}
    {\includegraphics[width=1in]{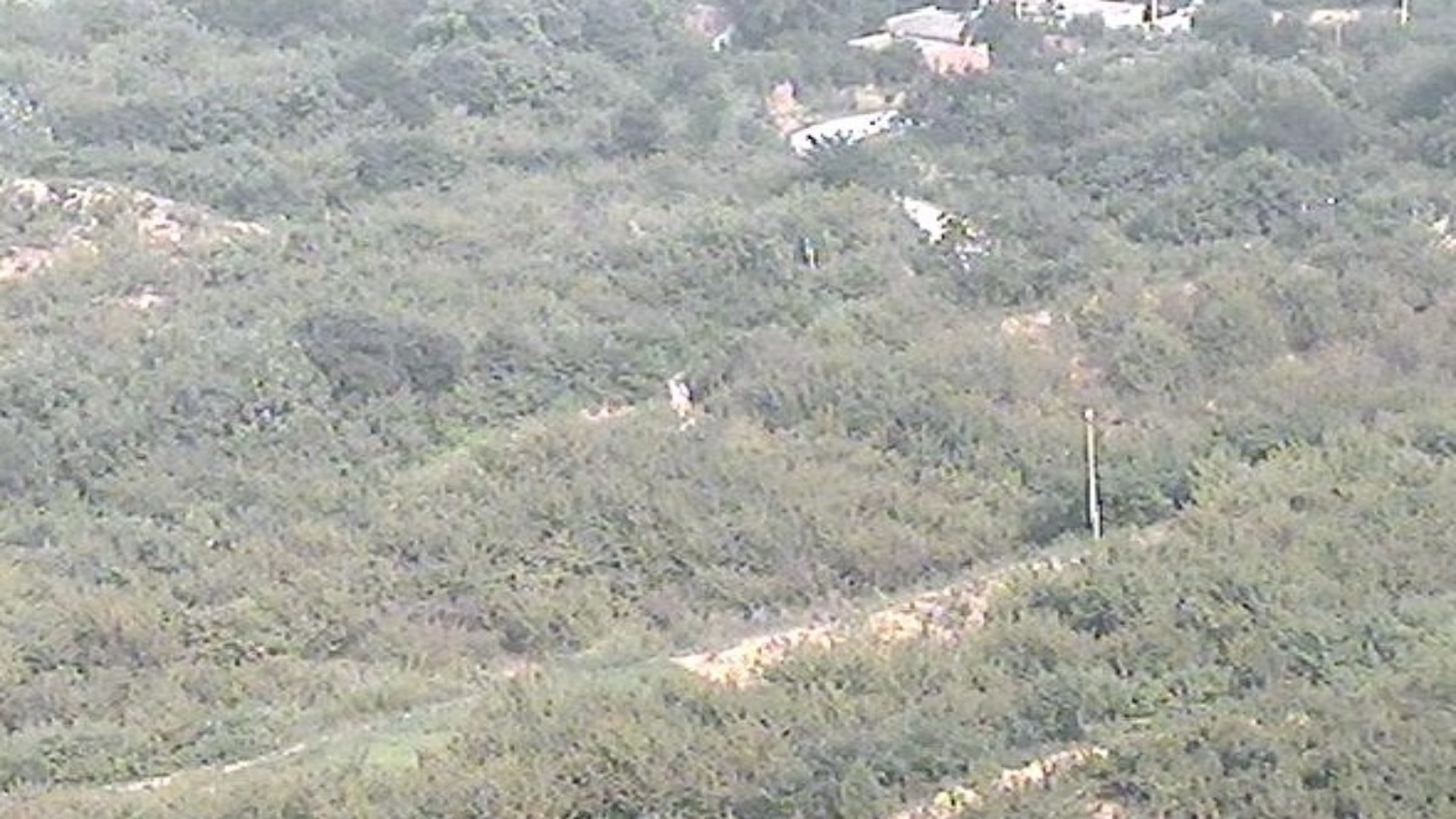}}
    {\includegraphics[width=1in]{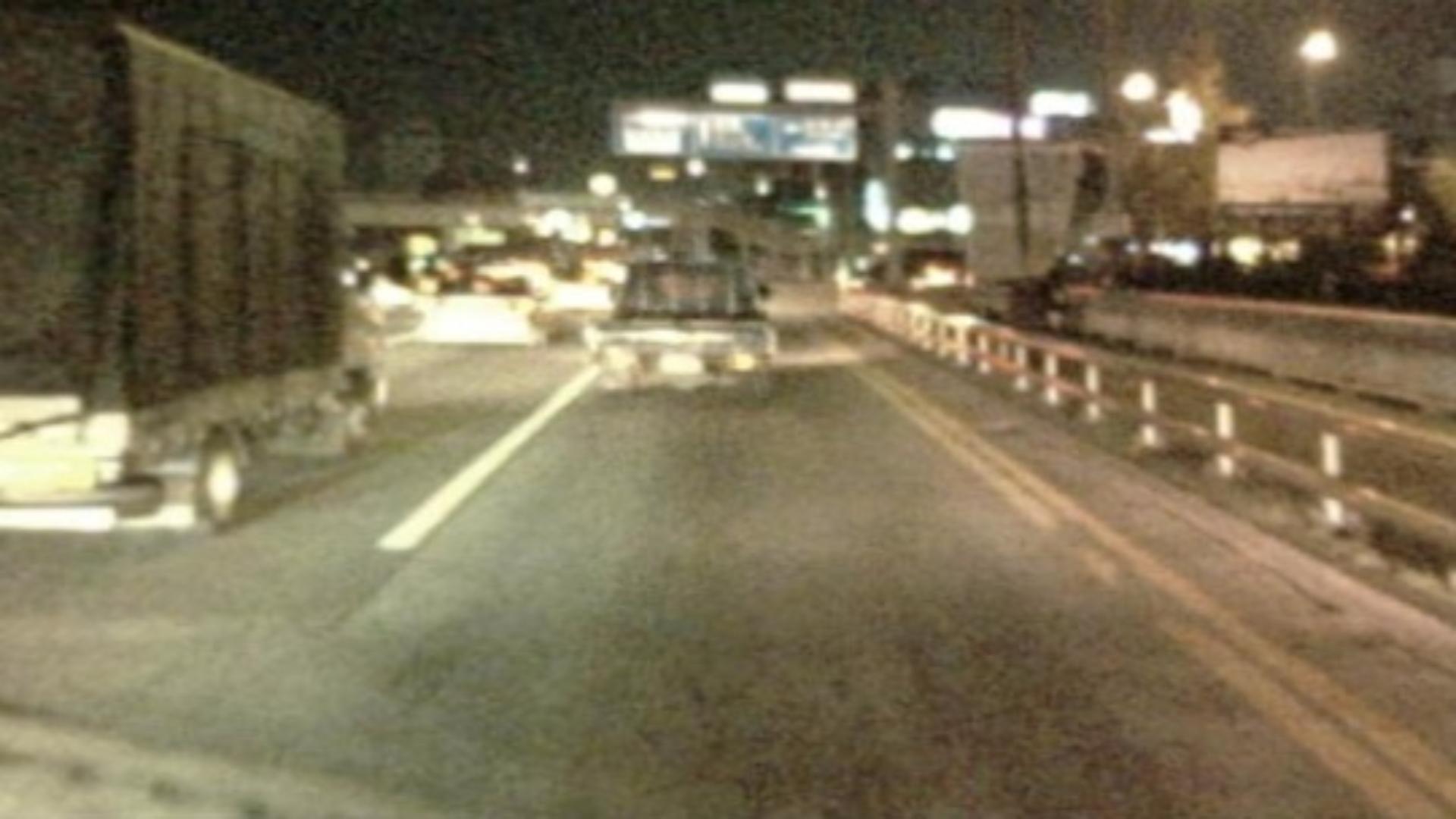}}
    {\includegraphics[width=1in]{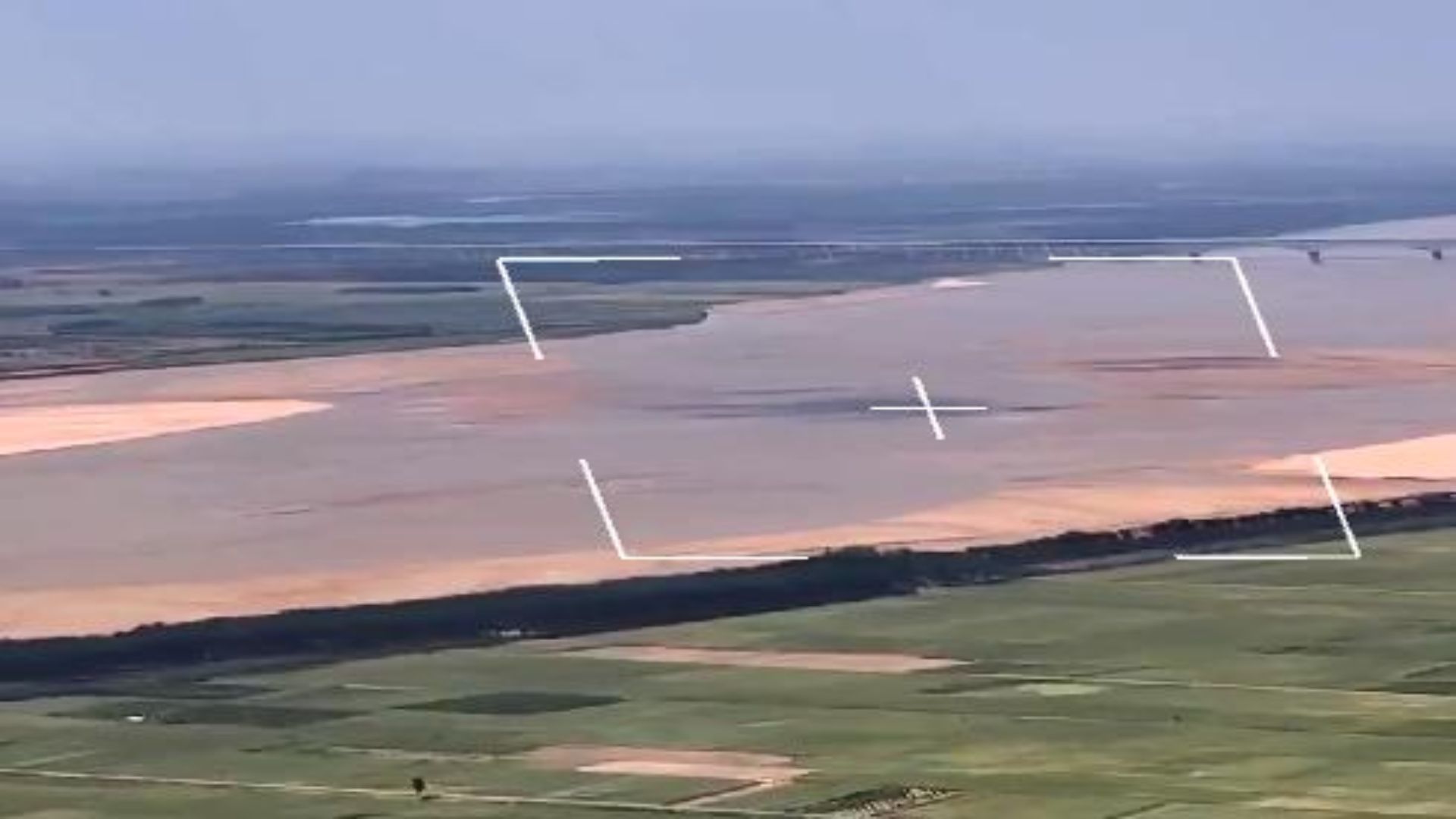}}

    \subfigure[Asteroid]{\includegraphics[width=1in]{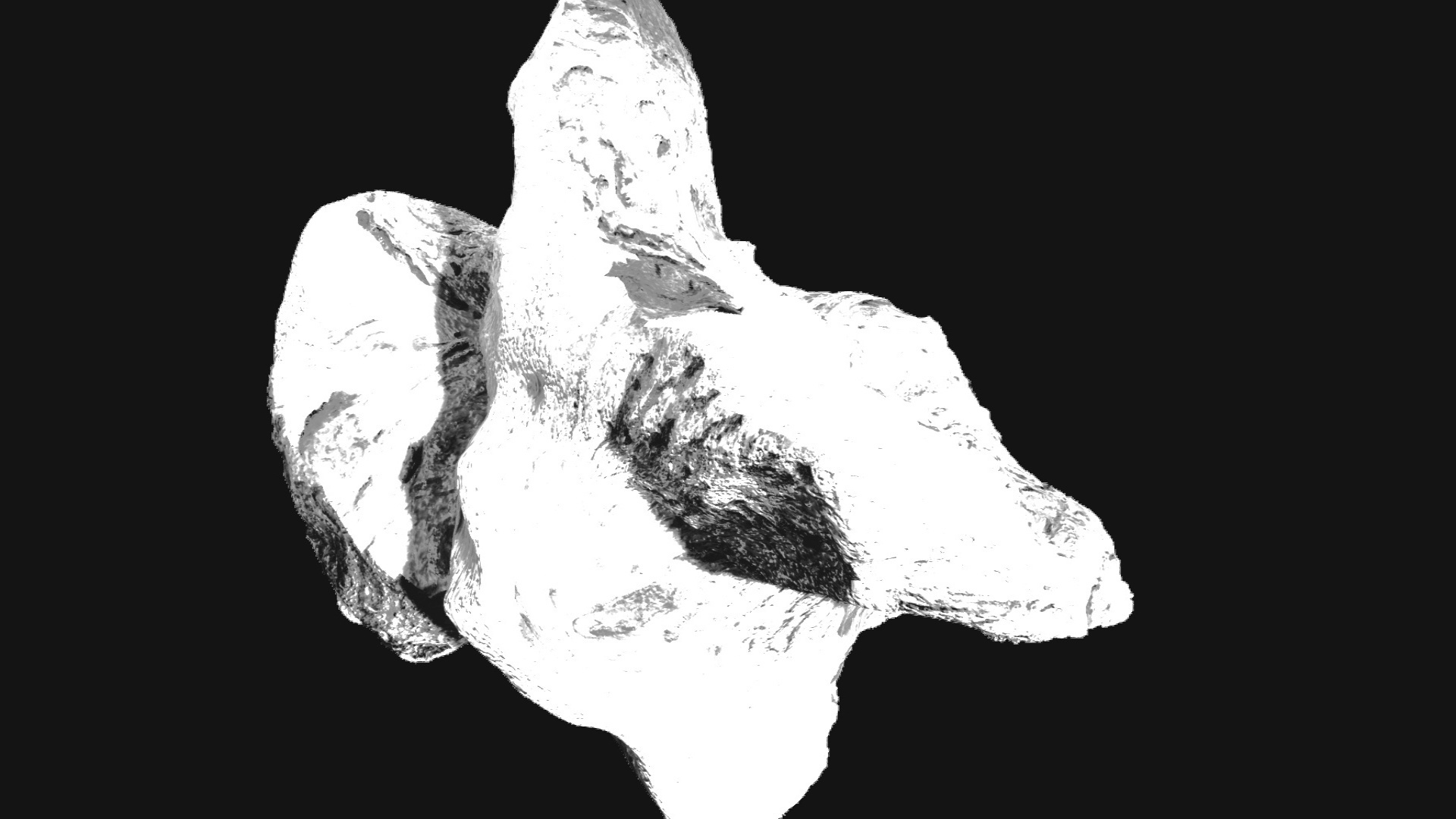}}
    \subfigure[Building]{\includegraphics[width=1in]{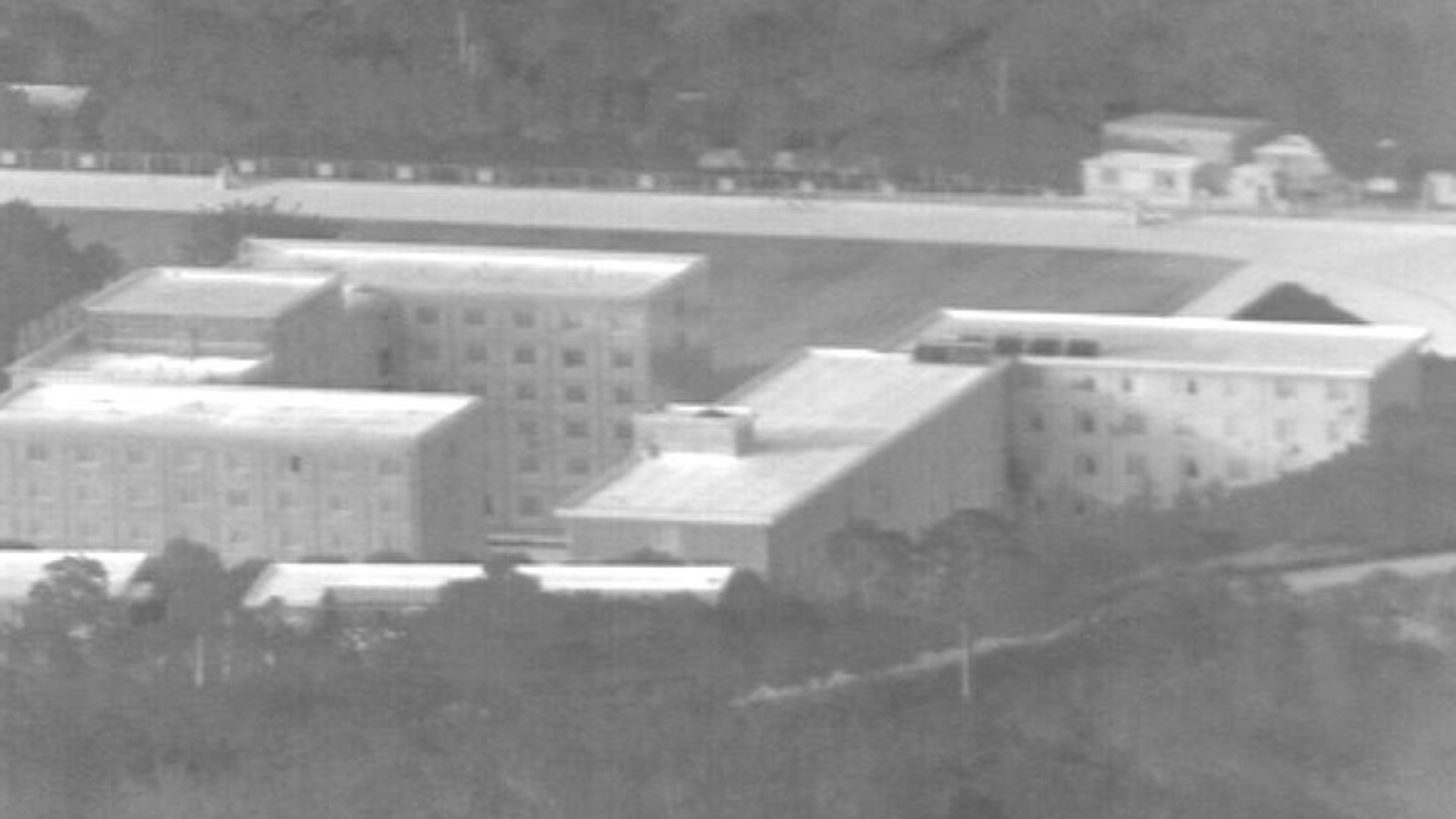}}
    \subfigure[Country]{\includegraphics[width=1in]{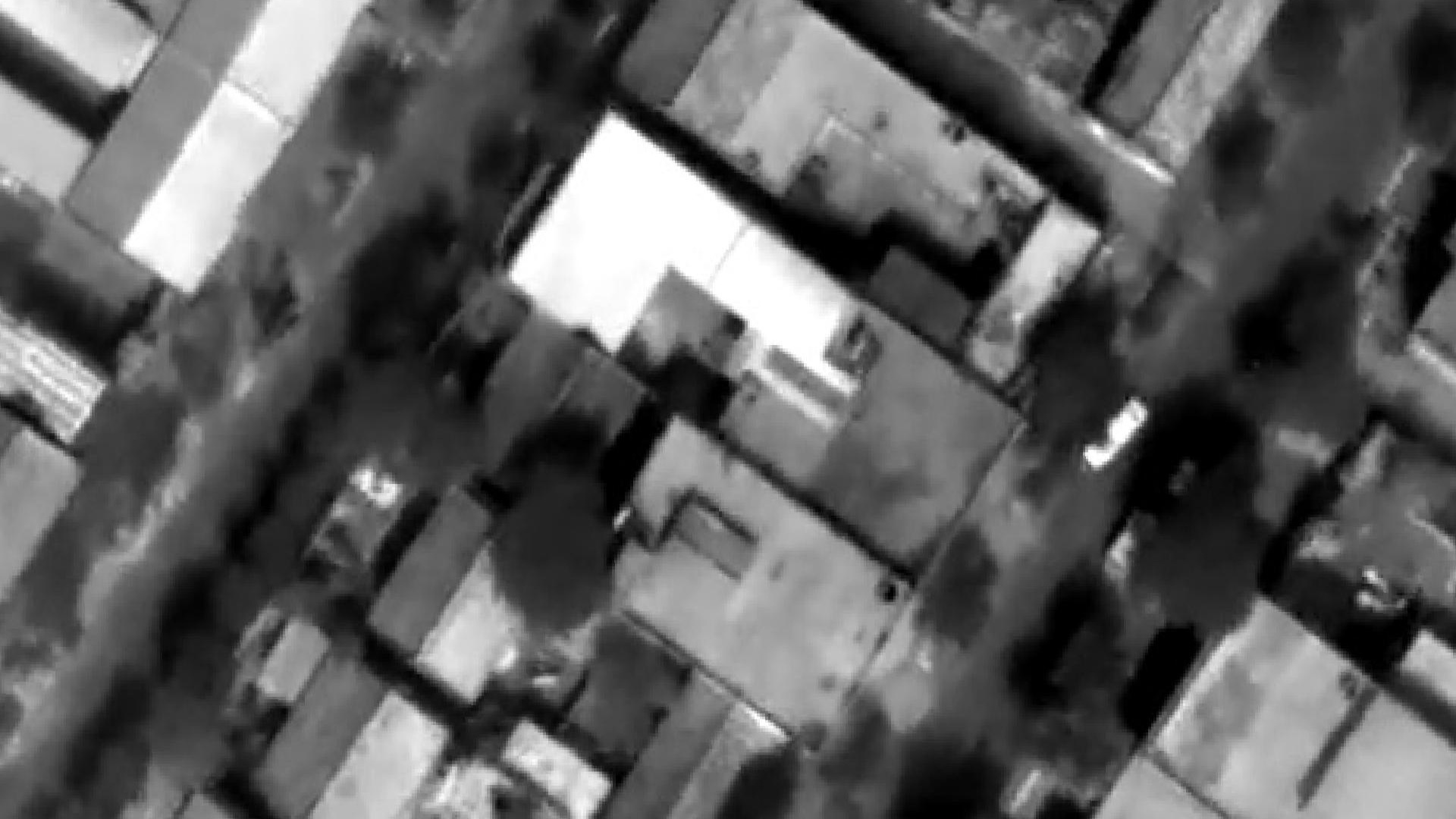}}
    \subfigure[Field]{\includegraphics[width=1in]{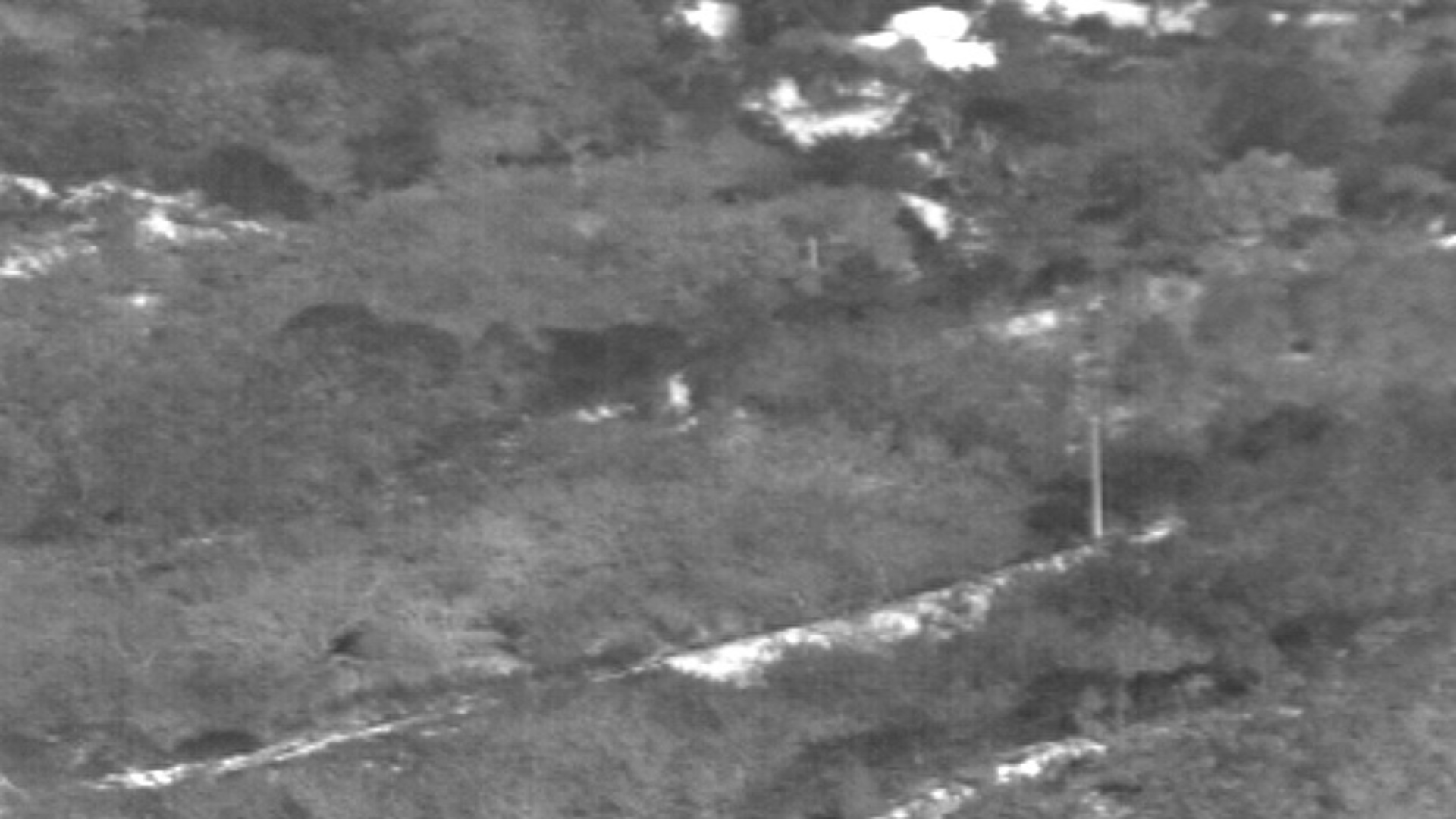}}
    \subfigure[Street]{\includegraphics[width=1in]{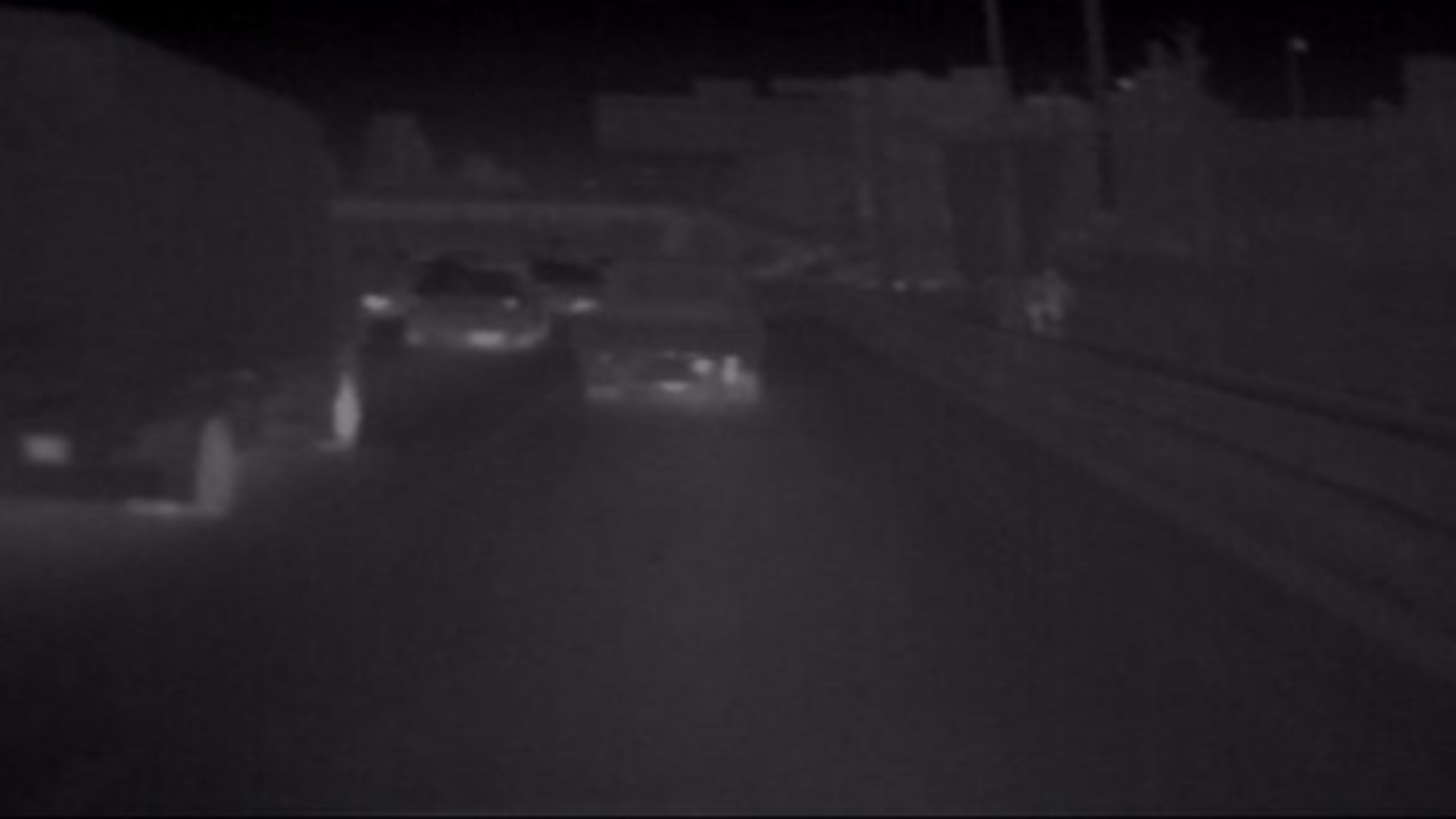}}
    \subfigure[Water]{\includegraphics[width=1in]{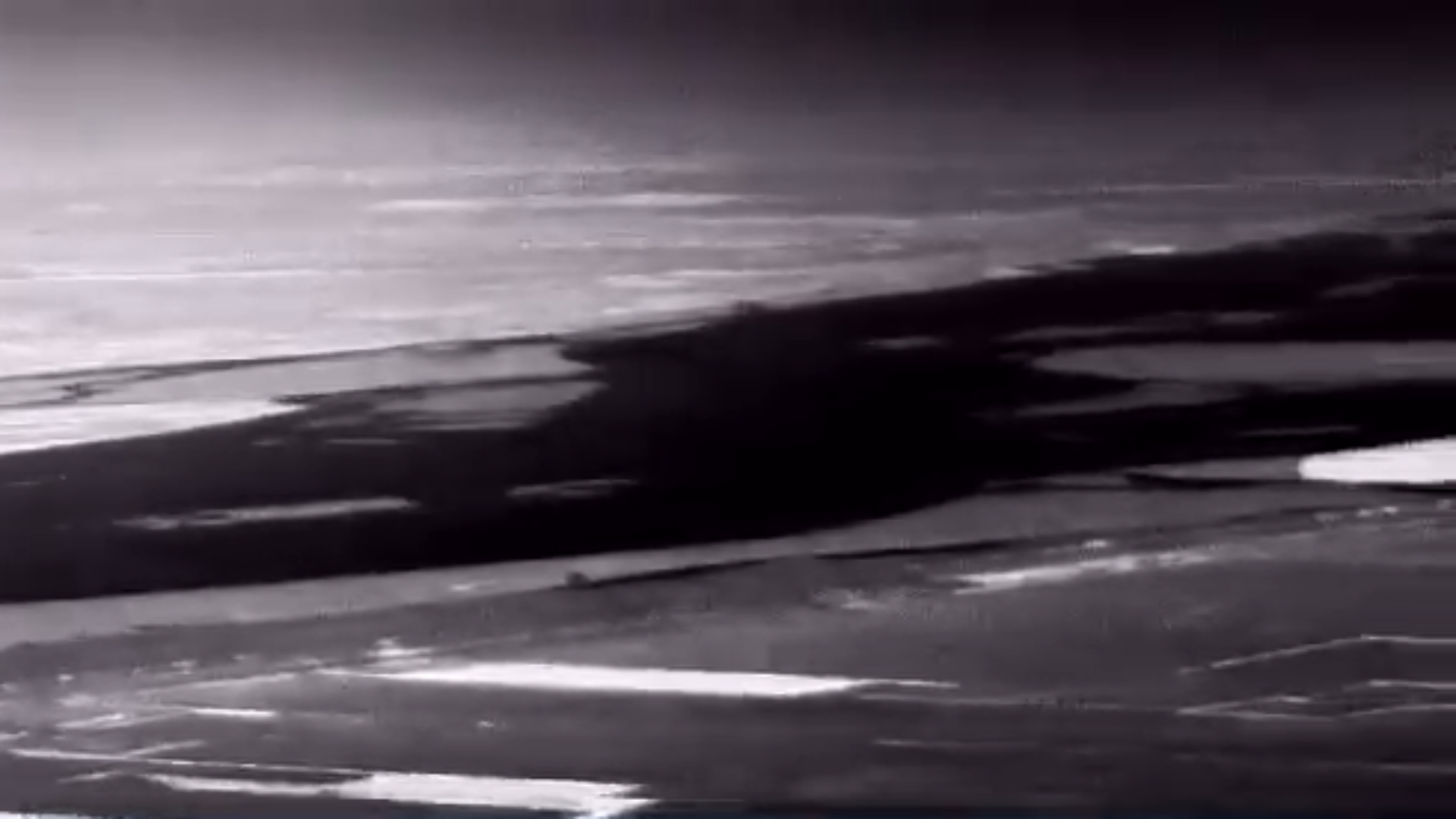}}

    \caption{Cross-spectral pairs samples from our VL-CMIM dataset. The first row corresponds to VIS images and the second row to LWIR images.}
    \label{data}
\end{figure*}

\begin{table*}[ht]
    \renewcommand\arraystretch{1.5}
    \centering
    \caption{Number of image pairs and patch pairs for each category in VL-CMIM dataset}
    \scalebox{1}
    {\begin{tabular}{c c c c c c c}
     \hline
     Category & Asteroid & Building & Country & Field & Street & Water \\
     \hline
  No. of Image Pairs  & 219 & 316 & 207 & 201 & 203 & 154\\
    \hline    
  No. of Patch Pairs & 784558 & 845188 &	324930 &	695742 & 343190 & 239051\\
     \hline
    \end{tabular}
     \label{tab:vis-lwir}}
\end{table*}

\textbf{Image Registration.}  Although visible light images and long-wave infrared images are shot by a binocular camera, they are not aligned due to the different field sizes of different sensor cameras. We clipped and registered visible-infrared image pairs so that they have exactly the same field of vision and the same image size. For this multi-modal image registration task, it is difficult to just apply automatic detection registration methods, so we chose a semi-manual method. We select alignment points in two images and find the transformation parameters to align them, typically using projection or affine transformations. However, images obtained from the simulation software are already synchronized in both time and space, eliminating the need for manual alignment. Fig. \ref{reg} shows the comparison of visible and infrared images before and after registration.

\begin{figure}[ht]
    \centering
    {
        \includegraphics[width=0.3\textwidth]{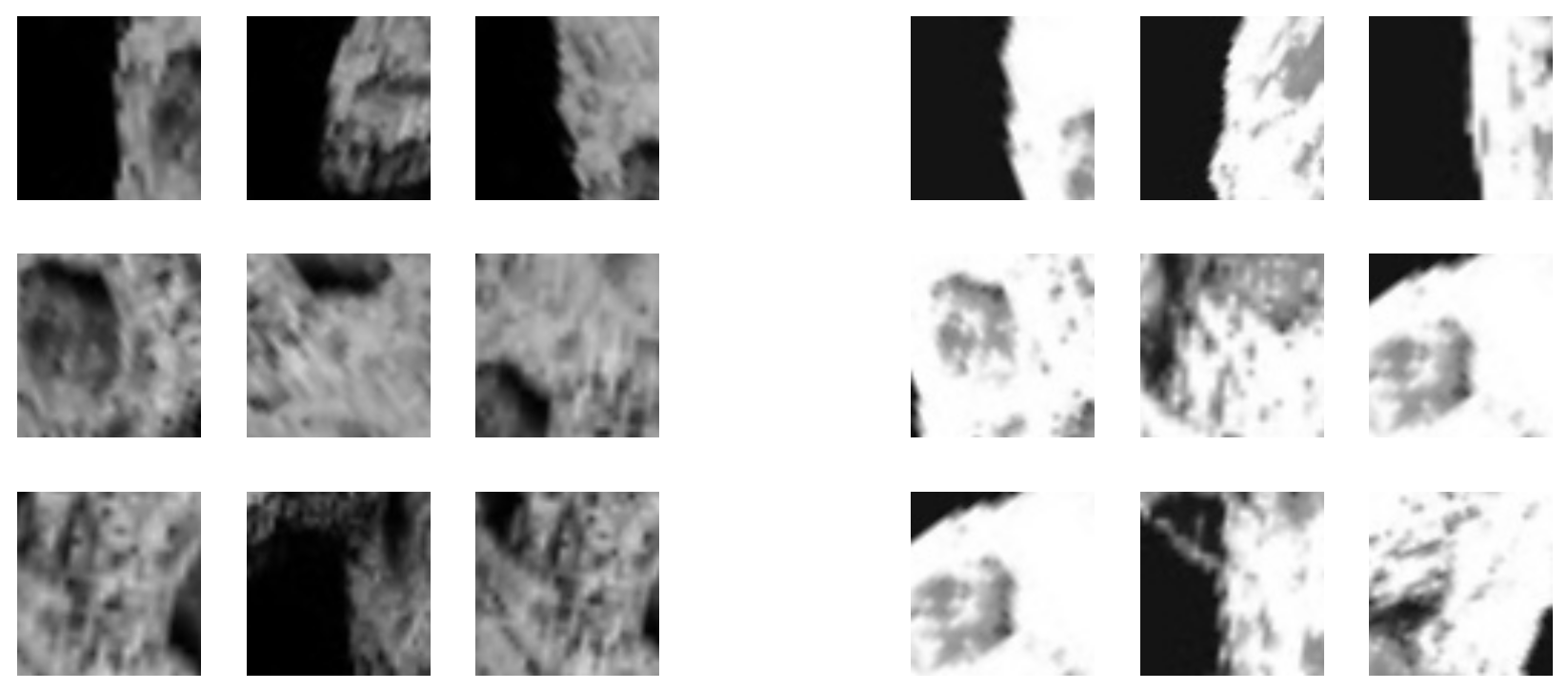}
        \label{img1}
    }
    \caption{Image patches from the training set. The left corresponds to visual grayscale images and the right to LWIR images. }
    \label{patch}
\end{figure}

\textbf{Generated Patches.}  Similar to \cite{quan2021multi,quan2019afd,moreshet2021paying,baruch2021joint,mishchuk2017working}, we construct visible and long-wave infrared image patch matching dataset based on VL-CMIM dataset. In the VL-CMIM dataset, the image exists as a set of alignment pairs, that is, one image for each mode, as shown in Fig. \ref{patch}. In this section, ORB operator is adopted to detect keypoints in visible and long-wave infrared images. Each feature point represents a pixel position and stores information about keypoint coordinates and direction. For each keypoint, sample pairs of size 64 $\times$ 64 pixels are extracted. Positive sample pairs are obtained based on the coordinate positions of corresponding keypoints, while negative sample pairs are randomly selected from other patches. The ratio of positive to negative samples is maintained at 1:1. This sampling approach ensures a balanced dataset for training.

Finally, our VL-CMIM dataset consists of 2600 images in 6 categories, i.e., asteroid, build, country, field, street and water, as shown in Fig. \ref{data} and Table \ref{tab:vis-lwir}.

\begin{figure*}[ht]
    \centering
    \includegraphics[width=1\textwidth]{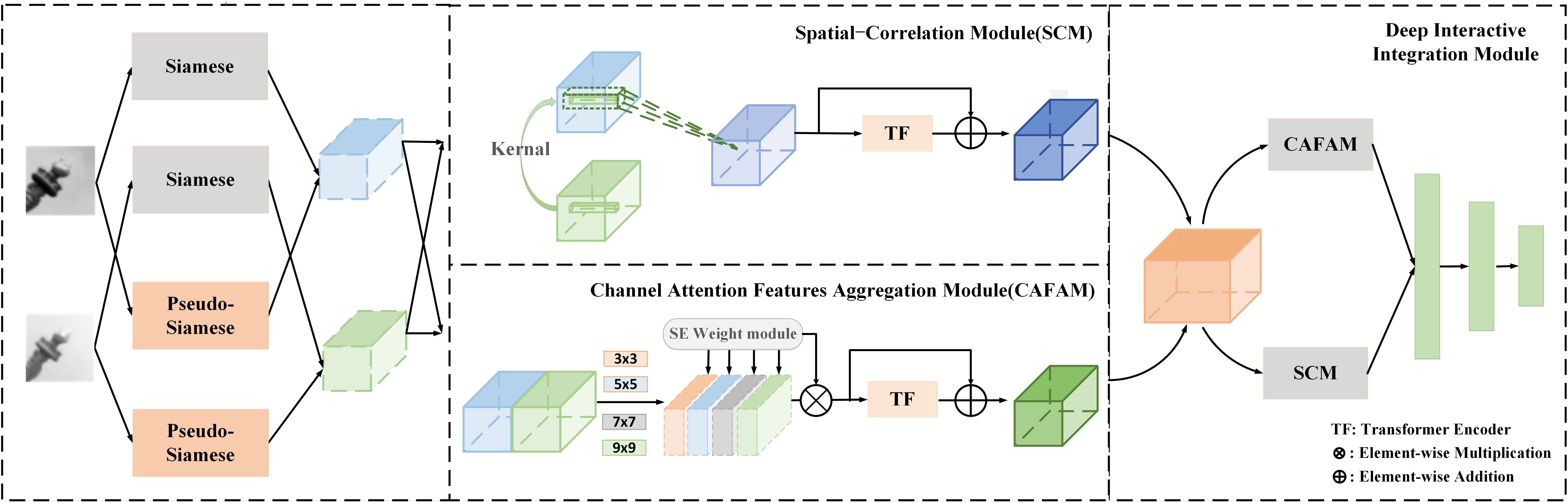}
    \caption{Overview of the pipeline of our proposed framework for image matching. Firstly, the four-branch multi-modal feature extraction module extracts multi-modal features from image patches. Secondly, the consistency features and discriminant features learned from the network are fused and passed through the spatial correlation feature extraction module and the multi-scale channel attention feature aggregation module respectively. Finally, the deep interactive fusion module is used to splice the captured features at the deep level, and interact again in space and channel. }
    \label{DeepCorrNet}
\end{figure*}

\section{Methodology} \label{scca}


In this section, we will first give an overview of our proposed DPI-QNet. Then, we will introduce each module in a more detailed way. As shown in Fig .\ref{DeepCorrNet}, our model contains four main parts, i.e., Four-branch Multi-modal Feature Extraction, Spatial Correlation, Multi-scale Channel Attention Feature Aggregation, and Deep Interactive Fusion. We will introduce these modules in Section \ref{0}, \ref{1}, \ref{2} and \ref{3}, respectively. 


\subsection{Overview} \label{Overview}


Given two source image patches, we first employ four branches to learn the correspondence between multi-modal image pairs. Then, the consistency features and discriminant features learned from the module were fused and fed into the spatial correlation feature extraction module and the multi-scale channel attention feature aggregation module respectively. The former uses the correlation layer \cite{dosovitskiy2015flownet} to combine the two branches and calculate the multiplicative sum of the pixel values corresponding to the image. The latter is designed to learn more multi-scale feature representation through an efficient pyramid attention segmentation module, and adaptively re-calibrate the multi-dimensional channel attention weight.
After that, the deep interactive fusion module is used to splice the captured features at the deep level, and interact again in space and channel.


\subsection{Four-branch Multi-modal Feature Extraction}\label{0}

As previous research has shown that four-branch network architecture can significantly improve performance in multi-modal image matching tasks\cite{aguilera2016learning, en2018ts}, we use the four-branch network structure in the feature extraction part considering that the four-branch network structure can also better extract similar features and discriminative features in multi-modal images. As shown in the left part of Fig .\ref{DeepCorrNet}, it contains four branches with the same structure. The Siamese sub-network contains two feature extraction branches with the same structure and shared weights, used to encode features unrelated to the imaging modality. In contrast, the Pseudo-Siamese sub-network also has two feature extraction branches with the same structure but distinct weights, specifically encoding imaging modality-related image pair features. 

For each branch, it consists of six convolution layers, whose parameters are shown in Table \ref{tab:backbone}. Specially, an instance normalization is added before Batch Normalization for the first three convolution layers to help reduce the feature difference caused by the illumination variation and different imaging mechanisms.


\subsection{Spatial Correlation Module} \label{1}





Inspired to study\cite{dosovitskiy2015flownet}, this section proposes a feature augmentation module based on spatial correlation. The spatial correlation module is shown in Fig .\ref{DeepCorrNet}.

The module consists of two steps.
 First, the correlation layer is utilized to merge two feature maps and learn the degree of correlation between them. Specifically, the second feature map is correlated with the first feature map. This correlation process helps establish the relationship between different features and enables the model to understand how they are related to each other.
Next, a Transformer encoder architecture is employed to establish long-range dependencies and capture global context information. The Transformer encoder structure is added to the high-level feature map of the spatial correlation feature extraction module. By leveraging the self-attention mechanism, the model can enhance its ability to perceive global relationships and obtain rich contextual information.
 In order to improve image matching performance, as with Moreshet et al. \cite{moreshet2021paying}, load the VIT pre-training model during training.




The calculation of the correlation layer is similar to that of the convolutional layer, with the main difference being that the correlation layer performs element-wise multiplication and summation instead of learning feature weights like the convolutional layer. The specific definitions are as follows:
\begin{equation}
    \begin{split}
        c(\mathbf{x}_1, \mathbf{x}2) &= \sum_{\mathbf{o} \in \boldsymbol{\Omega}} [{f}_1(\mathbf{x}_1 + \mathbf{o}) \otimes {f}_2(\mathbf{x}_2 + \mathbf{o})] \\
        \boldsymbol{\Omega} &= [-k, k] * [-k, k]
    \end{split}
\end{equation}

Here, $\mathbf{f}_1$ and $\mathbf{f}_2$ represent the first and second feature maps, respectively. $\mathbf{x}_1$ and $\mathbf{x}_2$ denote the corresponding positions in the two feature maps. $k$ represents the size of the comparison area. The correlation layer calculates the correlation between local patches of the two feature maps by performing element-wise multiplication followed by summation over the defined comparison area $\boldsymbol{\Omega}$.



Consider the computational complexity of attention operators limits the low resolution of inputs in Transformer.  As a result, in this module, Transformer encoder structure is added to the high-level feature map of the spatial correlation feature extraction module, and the self-attention mechanism is used to increase the global sensitivity field and obtain global information and rich context information. 

\subsection{Multi-scale Channel Attention Feature Aggregation Module} \label{2}

Multi-scale feature integration strategy and proper attention mechanism are proved to be beneficial for increasing the accuracy of metric learning \cite{hou2021coordinate,zhang2021epsanet}. Inspired by this fact, we propose a {Multi-scale Channel Attention Feature Aggregation Module (CAFAM), which encodes the similarity of different scale features of the input image patch pair respectively, and then integrates them together to increase the accuracy of metric learning. 

As illustrated in the middle part of Fig. \ref{DeepCorrNet}, the CAFAM module mainly contains three steps. Firstly, four convolution layers with different receptive fields (3 $\times$ 3, 5 $\times$ 5, 7 $\times$ 7, and 9 $\times$ 9) are utilized to generate four feature map groups. By splitting the input feature maps into four groups with different scales, CAFAM module can better measure the similarity of each scale. Then, Squeeze-and-Excitation attention (SE)\cite{hu2018squeeze} is  performed sequentially to encode  channel-wise correlation for each feature map group. Finally, the four groups of feature map refined by Transformer encoder are regarded as the outputs of the CAFAM module. 


Specifically, given an input feature map $ \mathbf{F}\in R^{L\times H \times W} $, the output of CAFAM module represented by $\mathbf{F}^{'}\in R^{L \times H \times W}$ can be computed as:

\begin{equation} \label{flow}
	\begin{split}
            \mathbf{F_0, F_1, F_2, F_3} &= \alpha (f^{3 \times 3}(\mathbf{F}), f^{5 \times 5}(\mathbf{F}), f^{7 \times 7}(\mathbf{F}), f^{9 \times 9}(\mathbf{F})) \\
            \mathbf{F_{i}^{'}} &= (SE(\mathbf{F_i})) \otimes \mathbf{F_i} \\
            \mathbf{F^{'}} &= TF(Concat(\mathbf{F_{0}^{'}}, \mathbf{F_{1}^{'}}, \mathbf{F_{2}^{'}}, \mathbf{F_{3}^{'}}))
	\end{split}
\end{equation}
where $f^{n\times n}$ represents the  convolution layer with kernel size of $n \times n$. $\mathbf{F_i} \in R^{C \times H \times W} $ ( $i=1, 2, 3, 4; C=L/4$) denotes one of four different scale feature maps.  $\alpha$ represents Sigmoid activation function, $\otimes$ is an element-wise multiplication. $SE(\cdot)$ is the Squeeze-and-Excitation attention, TF is Transformer encoder operation. $\mathbf{F_i^{'}} \in R^{C \times H \times W} $ denotes multi-scale feature map.

Squeeze-and-Excitation can encode the relationship among feature channels by an attention vector, which is calculated among different channels of feature maps. The details about our implementation of SE model are as follows. The input feature maps firstly go through a global pooling layer, and output a vector with the same size as the number of input feature map channels. Then, a fully connected layer with 32 units followed by a ReLU activation function, a fully connected layer with $C$ (the channel size of the input feature map) units,  and a Sigmoid function are performed to generate the attention vector. Finally the input feature maps are weighted by the attention vector, and element-wise added with themselves to produce the channel-wise attentive features. Through the SE module, the feature maps contributing to the matching  task are emphasized, and the others are restrained.

\begin{figure*}[ht]
    \centering
    \subfigure[]{\includegraphics[width=2.5in]{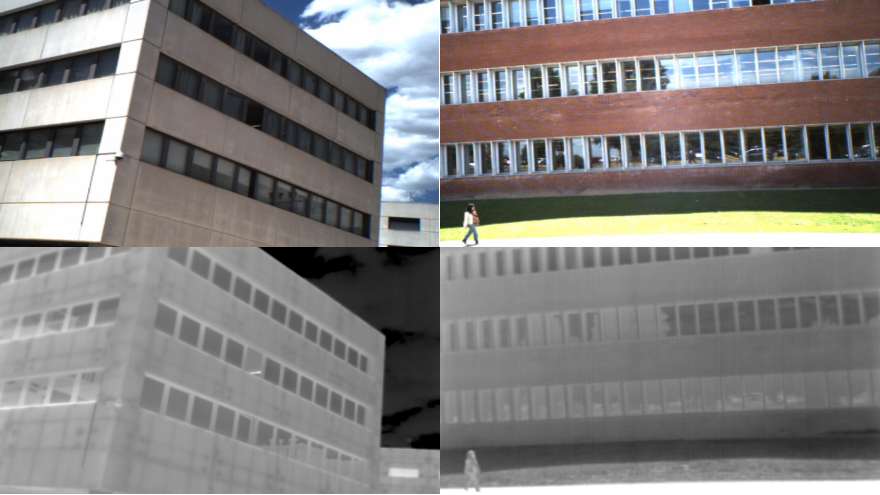}}
    \subfigure[]{\includegraphics[width=2.1in]{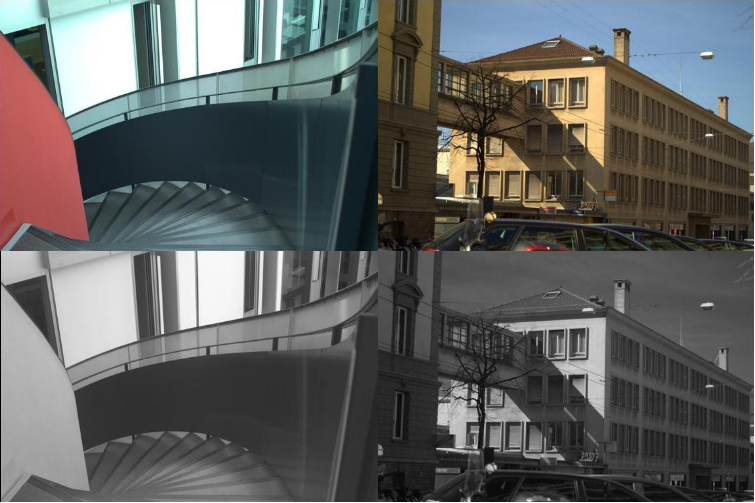}}
    \subfigure[]{\includegraphics[width=1.4in]{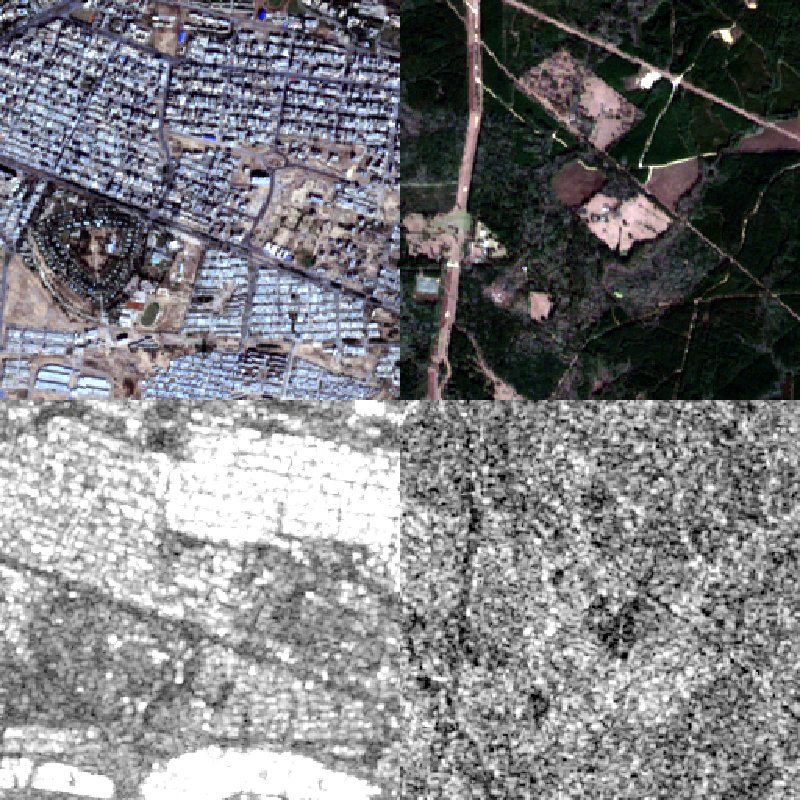}}
     \subfigure[]{\includegraphics[width=6in]{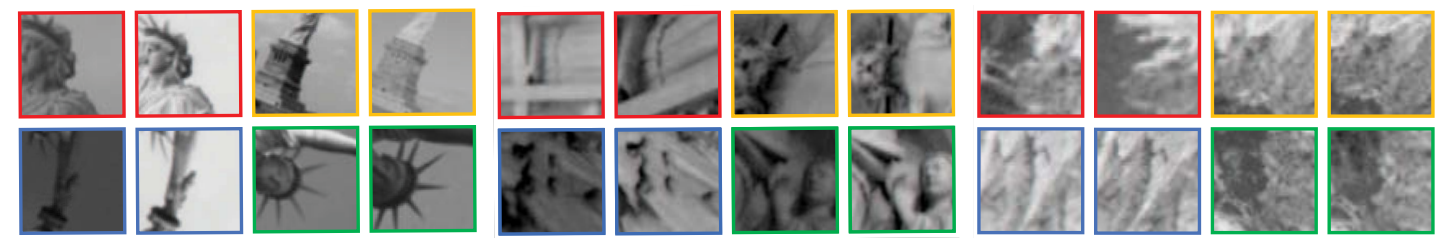}}
    \caption{Sample images of datasets used. (a) VIS-LWIR patch dataset. (b) VIS-NIR patch dataset. (c) Optical-Sar patch dataset. (d) Brown patch dataset. }
    \label{data_e}
\end{figure*}

\subsection{Deep Interactive Fusion Module} \label{3}
Deep Interactive feature fusion is very important for accurate prediction, which is denoted $DIFP$.
Based on spatial correlation feature extraction module and multi-scale channel attention feature aggregation module, the output feature space information is rich and can retain more feature details. However, in order to enrich the feature semantic information learned by the network and cover a larger space area, this chapter introduces the deep interactive fusion strategy. 

As depicted in the right part of Figure \ref{DeepCorrNet}, the proposed architecture incorporates a connection between the output features of the spatial correlation module and the multi-scale channel attention feature aggregation module. This connection allows for the flow of information between these modules, facilitating the transmission of cross-domain consistency features between the two modes.
The connected features are then passed through the respective modules, promoting the exchange of information and fostering the integration of features from both modes. This enables the models to benefit from the complementary information present in each mode.
Finally, three fully connected layers with sizes of 512, 128, and 2 are utilized to predict the final network results. These fully connected layers map the integrated features to the desired output dimension, enabling the network to make predictions based on the learned representations.

By combining the information from the spatial correlation module, the multi-scale channel attention feature aggregation module, and the cross-domain consistency features, the network can leverage the strengths of each module and make accurate predictions for the given task.

\section{Experiments} \label{exper}

In this section, we first introduce the dataset used for the experiments. The implementation details of the training process are then introduced. Then, a series of ablation and compare experiments are described. They were performed on the VIS-LWIR, VIS-NIR and Brown dataset. Eventually, comparison results with some state-of-the-art methods are exhibited.

\subsection{Implementation Details }
1) Experimental Environment and Parameter Settings: In training processing, our model is implemented by Pytorch with Adam optimizer with a learning rate of 0.0001. The batch size is 128. The training time is set to 100 epochs, the momentum is initially set to 0.9 with the decay factor 0.9. 
\begin{table}[h]
    \centering
    \scalebox{0.9}{
        \begin{tabular}{*{6}{c}}
            \toprule
            Layer & Output & Kernel Size & Stride & Padding  \\
            \midrule
            Conv0 & $64\times64\times32$ & $3\times3$ & 1 & 1   \\
            Conv1 & $64\times64\times32$ & $3\times3$ & 1 & 1   \\
            Conv2 & $32\times32\times64$ & $3\times3$ & 1 & 1   \\
            Conv3 & $32\times32\times64$ & $3\times3$ & 1 & 1   \\
            Conv4 & $32\times32\times128$ & $3\times3$ & 1 & 1  \\
            Conv5 & $16\times16\times128$ & $3\times3$ & 1 & 1  \\
            Conv6 & $16\times16\times128$ & $3\times3$ & 1 & 1  \\
            Conv7 & $16\times16\times128$ & $3\times3$ & 1 & 1  \\     
            \bottomrule
        \end{tabular}}
    \caption{The Siamese and Pseudo-Siamese backbone CNN architecture}
    \label{tab:backbone}
\end{table}


2) Evaluation Metric: The FPR95 is often used as the image patch matching task evaluation metric to quantitatively evaluate the matching performance of the network. Specifically, it means that the false positive rate at true positive rate (positive recall) equal to 95\%(FPR95). The metrics are defined as follows:
\begin{equation}
    FPR95=\frac{FP}{FP+TN},  TP >= 95\%
\end{equation}
where TP represents the number of correct matches, FP represents the number of false matches in the 95\% set. TN represents the correct match in the remaining 5\%. The smaller the FPR95 value, the better the matching performance.

\subsection{Datasets Description}

In order to confirm the effectiveness of our proposed method, we carry out the experiments on VIS-NIR, VIS-LWIR, Optical-SAR, VIS-LWIR dataset and Brown dataset as shown in Fig. \ref{data_e}.

VIS-NIR also known as Nirscene\cite{brown2011multi}, contains visible spectrum and near-infrared spectrum. It is a standard multi-modal image matching dataset \cite{quan2019afd}. The dataset contains 477 pairs of images in 9 categories, i.e., Country, Field, Forest, Indoor, Mountain, Oldbuilding, Street, Urban and Water. 
Like the methods \cite{baruch2021joint, moreshet2021paying, quan2021multi}, the proposed model is trained on the category Country and test on the other categories. Since great differences among these categories as shown in Fig. \ref{vis-nir}., it is hard to obtain a good generalization performance on all test categories.

VIS-LWIR\cite{aguilera2015lghd} is a cross-spectral dataset of visible spectrum and long-wave infrared spectrum. Compared with near-infrared images, the appearance difference of long-wave infrared image is larger and the matching difficulty is higher. VIS-LWIR contains 44 visible and thermal infrared image pairs, all of which are strictly aligned in time and space.
Similar to VIS-NIR dataset, we also used SIFT detection sub-to construct cross-modal image patch matching dataset with image patch resolution of 64 $\times$ 64 pixels. Similar to previous studies, the following two image patch matching networks were trained on half of the image patch pairs and tested on the other half of the image patch pairs.


Optical-SAR is a multimodal image patch dataset containing optical images and synthetic aperture radar images. Optical image can provide more scene information, but it is greatly affected by illumination. Synthetic aperture radar image can work all day, all weather, widely used in agriculture, military and other fields. The matching of optical image and SAR image is a key task in the application of multimode remote sensing image and has important significance. SEN1-2\cite{schmitt2019sen12ms, SEN1-2} is the first large open source dataset in the field of multi-sensor data fusion, containing a total of 282,384 pairs of optical images and corresponding SAR images from four different seasons, with a resolution of 256 $\times$ 256 for both modals. Similar to VIS-NIR dataset, image patch pairs were constructed from SEN1-2 dataset in the same way for follow-up studies. 
A total of 583,180 image block pairs were obtained for training and 248,274 pairs were used as test.

Brown dataset also referred to the Multi-view Stereo Correspondence dataset\cite{brown2010discriminative}, which is a single spectural image dataset, and is a benchmark in image patch matching task. 
It consists of corresponding patches sampled from 3-D reconstructions. The Brown dataset is composed of three subset: Liberty, Notredame and Yosemite. Each subset contains 100K, 200K, and 500K image patch pairs. The patch size is 64 $\times$ 64. Half of the patches are matching pairs having the same 3-D points ID, and the corresponding interest points are within 5 pixels in position, 0.25 octaves of scale, and ($\frac{\pi}{8}$) radians in angle. Half are non-matching pairs that have different 3-D points ID and correspond to interest points exceed 10 pixels in position, 0.5 octaves of scale, and ($\frac{\pi}{4}$) radians in angle. 
Following\cite{tian2017l2, han2015matchnet, zagoruyko2015learning}, we train on one of the three subset and test on the other subset. 
%

    

\begin{table}[ht]
    \renewcommand\arraystretch{1.3}
    \centering
    \caption{Comparisons with the-state-of-the-art on the VIS-LWIR dataset and OPTICAL-SAR dataset}
    
    \begin{tabular}{c | c c}
    \hline
    \diagbox{Method}{Dataset} & VIS-LWIR & Optical-SAR \\
        \hline
        Siamese \cite{simo2015discriminative} & 42.62 &  17.56 \\
        Pseudo-Siamese \cite{zagoruyko2015learning} & 43.27 & 19.30 \\
        2Channel \cite{zagoruyko2015learning} & 22.95 & 7.35\\
        Hybrid \cite{baruch2021joint}  & 18.09 & 14.90\\
        \textbf{DPI-QNet} & \textbf{1.97} & \textbf{0.78} \\
    \hline
    \end{tabular}
    \label{tab:scca-compare-vis-lwir}
\end{table}

\begin{table*}[ht]
    \renewcommand\arraystretch{1.3}
    \centering
    \caption{Comparisons with the-state-of-the-art on the VIS-NIR image dataset}
     \scalebox{0.98}{
    \begin{tabular}{c c c c c c c c c c}
        \hline
        Method & Field & Forest & Indoor & Mountain & Oldbuilding & Street & Urban & Water & Mean \\
    
         \hline
        SIFT \cite{lowe2004distinctive} & 39.44 & 11.39 & 10.13 & 28.63 & 19.69 & 31.14 & 10.85 & 40.33 & 23.95 \\
        GISIFT \cite{firmenichy2011multispectral} & 34.75 & 16.63 &	10.63 &	19.52 &	12.54 &	21.80 &	7.21 &	25.78 &	18.60 \\
        EHD \cite{aguilera2012multispectral} & 33.85 &	19.61 &	24.23 &	26.32 &	17.11 &	22.31 &	3.77 &	19.80 &	20.87 \\
        LGHD \cite{shechtman2007matching} & 16.52 &	3.78 &	7.91 &	10.66 &	7.91 &	6.55 &	7.21 &	12.76 &	9.16 \\
        PN-Net \cite{balntas2016pn} & 20.09 & 3.27 & 6.36 &	11.53 &	5.19 &	5.62 &	3.31 &	10.72 &	8.26 \\
        Q-Net \cite{savinov2017quad} & 17.01 &	2.70 &	6.16 &	9.61 &	4.61 &	3.99 &	2.83 &	8.44 &	6.91 \\
        L2-Net \cite{tian2017l2} & 16.77 &	0.76 &	2.07 &	5.98 &	1.89 &	2.83 &	0.62 &	11.11 &	5.25 \\
        HardNet \cite{mishchuk2017working} & 10.89 &	0.22 &	1.87 &	3.09 &	1.32 &	1.30 &	1.19 &	2.54 &	2.80 \\
        Siamese \cite{simo2015discriminative} & 15.79 &	10.76 &	11.60 &	11.15 &	5.27 &	7.51 &	4.60 &	10.21 &	9.61 \\
        Pseudo-Siamese \cite{zagoruyko2015learning} & 17.01 & 9.82 & 11.17 & 11.86 & 6.75 &	8.25 &	5.65 &	12.04 &	10.31 \\
        2-Channel \cite{zagoruyko2015learning} & 9.96 &	0.12 &	4.40 &	8.89 &	2.30 &	2.18 &	1.58 &	6.40 &	4.47 \\
        SCFDM \cite{quan2019cross} & 7.91 &	0.87 &	3.93 &	5.07 &	2.27 &	2.22 &	0.85 &	4.75 &	3.48 \\
        Hybrid \cite{baruch2021joint} & 5.62 &	0.53 &	3.58 &	3.51 &	2.23 &	1.82 &	1.90 &	3.05 &	2.52 \\
        Moreshet \& K+ \cite{moreshet2021paying} & 4.22 & 0.13 &	1.48 &	1.03 &	1.06 &	1.03 &	0.9	& 1.9 &	1.44 \\
        Quan \& W+ \cite{quan2021multi} & 4.21 &	0.11 &	1.12 &	0.87 &	0.67 &	0.56 &	0.43 &	1.90 &	1.23 \\
        AFD-Net \cite{quan2019afd} & 3.47 &	 0.08 &	1.48 &	0.68 &	0.71 &	 0.42 &	 0.29 &	1.48 &	1.08 \\
        \textbf{DPI-QNet} &  \textbf{1.80} &	0.84 &	 \textbf{0.58} & \textbf{0.31} & 0.58 &	0.81 &	0.47 &	2.18 &	0.95 \\
        \hline
    \end{tabular}}
    \label{tab:scca-compare-vis-nir}
\end{table*}

\begin{table}[ht]
    \renewcommand\arraystretch{1.3}
    \centering
    \caption{Comparisons with the-state-of-the-art on the VIS-LWIR dataset and OPTICAL-SAR dataset}
    
    \begin{tabular}{c | c c}
    \hline
    \backslashbox{Method}{Dataset} &  VIS-LWIR & Optical-SAR \\
        \hline
        Siamese \cite{simo2015discriminative} & 42.62 &  17.56 \\
        Pseudo-Siamese \cite{zagoruyko2015learning} & 43.27 & 19.30 \\
        2Channel \cite{zagoruyko2015learning} & 22.95 & 7.35\\
        Hybrid \cite{baruch2021joint}  & 18.09 & 14.90\\
        \textbf{DPI-QNet} & \textbf{1.97} & \textbf{0.78} \\
    \hline
    \end{tabular}
    \label{tab:scca-compare-vis-lwir}
\end{table}


\begin{table*}[ht]
    \renewcommand\arraystretch{1.3}
    \centering
    \caption{Comparisons with the-state-of-the-art on the Brown dataset}
    \scalebox{0.98}{
    \begin{tabular}{c c c c c c c c}
        \hline
        \multirow{1}*{Training} & \multirow{1}*{Notredame} & \multirow{1}*{Yosemite} & \multirow{1}*{Liberty} & \multirow{1}*{Yosemite} & \multirow{1}*{Liberty} & \multirow{1}*{Notredame}& \multirow{2}*{Mean}\\
        
        \cmidrule(lr){2-3}\cmidrule(lr){4-5}\cmidrule(lr){6-7}
        Test & \multicolumn{2}{c}{Liberty} & \multicolumn{2}{c}{Notredame} & \multicolumn{2}{c}{Yosemite}  \\
         
            \hline
             RootSIFT\cite{2012Three} & \multicolumn{2}{c}{29.65} & \multicolumn{2}{c}{22.06} & \multicolumn{2}{c}{26.71} & 26.14 \\
            L-BGM\cite{2012Learning} &	18.05 &	21.03 &	14.15 &	13.73 &	19.63 &	15.86 &	17.08 \\
            Convex optimization\cite{simonyan2014learning} & 12.42 &	14.58 &	7.22 & 6.82 &	11.18 &	10.08 &	10.38 \\
            TNet-TGLoss\cite{kumar2016learning} & 9.91 &	13.45 &	3.91 &	5.43 &	10.65 &	9.47 &	8.80 \\
            SNet-GLoss\cite{kumar2016learning} & 6.39 &	8.43 &	1.84 &	2.83 &	6.61 &	5.57 &	5.27 \\
            PN-Net\cite{balntas2016pn} & 8.13 &	9.65 &	3.71 &	4.23 &	8.99 &	7.21 &	6.98 \\
            Q-Net\cite{savinov2017quad} &	7.64 &	10.22 &	4.07 &	3.76 &	9.34 &	7.69 &	7.12 \\
            DeepDesc\cite{simo2015discriminative} & \multicolumn{2}{c}{10.90} & \multicolumn{2}{c}{4.40} & \multicolumn{2}{c}{5.69} & 6.99 \\			
            TFeat-ration\cite{balntas2016learning} &	8.07 &	9.53 &	3.47 &	4.23 &	8.53 &	7.24 &	6.84 \\
            TFeat-margin\cite{balntas2016learning} &	7.22 &	9.79 &	3.12 &	3.85 &	7.82 &	7.08 &	6.47 \\
            L2-Net\cite{tian2017l2} &	2.36 &	4.70 &	0.72 &	1.29 &	2.57 &	1.71 &	2.22 \\
            HardNet\cite{mishchuk2017working} &	1.49 &	2.51 &	0.53 &	0.78 &	1.96 &	1.84 &	1.51 \\
            MathchNet\cite{han2015matchnet} &	6.90 &	10.77 &	3.87 &	5.67 &	10.88 &	8.39 &	7.44 \\
            DeepCompare\cite{zagoruyko2015learning}  &	4.85 &	7.20 &	1.90 &	2.11 &	5.00 &	4.10 &	4.19 \\
            SCFDM\cite{quan2019cross} &	1.47 &	4.54 &	1.29 &	1.96 &	2.91 &	5.20 &	2.89 \\
            Quan \& W+ \cite{quan2021multi} &	1.47 &	2.09 &	0.50 &	0.77 &	1.69 &	1.75 &	1.38 \\
            Moreshet \& K+ \cite{moreshet2021paying} &	 0.35 &	0.91 &	1.31 &	0.85 &	1.58 &	0.41 &	0.9 \\
            AFD-Net \cite{quan2019afd} & 1.53 &	2.31 &	 0.47 &	0.72  & 1.63 &	1.88 &	1.42 \\
            MFD-Net \cite{yu2022multi} & 1.21 & 2.10 & 0.40 & 0.74 & 1.85 & 1.77 & 1.35\\

             \textbf{DPI-QNet} & 0.27 & 0.59 & 0.13 & 0.32 & 0.12 & 0.21 & 0.27\\
            \hline
            
    \end{tabular}}
    \label{tab:scca-compare-brown-1}
\end{table*}

\begin{table*}[ht]
    \renewcommand\arraystretch{1.3}
    \centering
    \caption{Ablation results evaluated using the VL-CMIM dataset.}
    \scalebox{0.98}{
\begin{tabular}{ c| c | c | c | c | c | c | c | c  c  c  c c c }
\hline
    Sia & Pre-Sia  & CL & EPSA & TF1 & TF2 & DI & $\mathbf{DPI-Q}$ & Asteroid & Field & Build & Street & Water & Mean  \\
    \hline
    \checkmark & \checkmark & & & & & & & 4.96 & 9.04 & 5.99 & 6.47 & 6.98 & 6.69  \\
   \checkmark& \checkmark& \checkmark  & & \checkmark & & & & 3.06 & 8.27 & 3.65 & 5.16 & 5.32 & 5.10 \\
    \checkmark & \checkmark &  & \checkmark & & \checkmark & & & 5.68 & 9.16 & 6.29 & 6.28 & 5.95 & 6.67 \\
    \checkmark & \checkmark & \checkmark  & \checkmark & & & & & 3.37 & 8.33 & 4.68 & 5.33 & 5.64 &5.47 \\
    \checkmark & \checkmark & & & \checkmark & \checkmark & & & 6.83 & 8.14 & 10.35 & 9.38 & 10.02 &8.94 \\
    \checkmark & \checkmark & \checkmark & \checkmark& \checkmark& \checkmark & & & 2.37 & 7.08 & 3.34 & \textbf{4.25} & 4.51 & 4.31 \\
    \checkmark & \checkmark &  &  &  &  & & \checkmark & 3.77 &7.77 & 5.24 & 5.96 & 6.03 &5.75\\
    \checkmark & \checkmark & \checkmark &\checkmark& \checkmark & \checkmark & \checkmark & & \textbf{2.02} & \textbf{5.35} & \textbf{3.38} & 4.75 &	\textbf{4.25} & \textbf{3.95} \\
    
     \hline
 \end{tabular}}
     \label{tab:ablation-scca}
\end{table*}


\begin{table*}[ht]
\renewcommand\arraystretch{1.3}
    \centering
    \caption{Cross-dataset Transfering Performance: trained on other datasets and test on Brown dataset}
    \scalebox{1.0}{
    \begin{tabular}{c | c c c | c}
    \hline
    \multicolumn{1}{c|}{\multirow{2}*{\diagbox{Train Dataset}{Test Dataset}}} &  \multicolumn{3}{c|}{Brown} & \multirow{2}*{Mean}\\
        \cline{2-4}
         & Notredame & Yosemite & Liberty \\
        \hline
        VIS-NIR & 2.43 & 3.10  & 2.13 & 2.55\\
        VIS-LWIR & 5.07 & 5.56 &  5.33 & 5.32 \\
        Optical-SAR & 15.94 & 16.86 & 15.87 & 16.22 \\
        VL-CMIM & 3.66 & 3.91 & 2.63 & 3.40 \\
        \hline
    \end{tabular}}
    \label{tab:Brown-transfer-learning}
\end{table*}

\begin{table*}[ht]
    \renewcommand\arraystretch{1.3}
    \centering
    \caption{Cross-dataset Transfering Performance: trained on other datasets and test on VIS-NIR dataset}
    \setlength{\tabcolsep}{1.5mm}{
    \scalebox{0.98}{
    \begin{tabular}{ l l |  c c c c c c c c  |c }
    \hline
         \multicolumn{2}{c|}{\multirow{2}*{\diagbox{Train Dataset}{Test Dataset}}} &  \multicolumn{8}{c|}{VIS-NIR} & \multirow{2}*{Mean}\\
        \cline{3-10}
        &  &  Field & Forest & Indoor & Mountain & Oldbuilding & Street & Urban & Water \\
        \hline
        \multirow{3}*{Brown} & Yosemite & 1.02 & 4.56 & 6.33 &  2.04& 2.55 & 2.13 & 1.01 & 1.02 & 2.61 \\
        & Notredame & 1.14 & 4.22 & 5.86 & 2.27 & 2.46 & 2.39 & 1.03 & 1.15 & 2.57 \\
        & Liberty & 4.79  & 3.74 & 3.05 & 1.64 & 1.59 & 3.60 & 1.62 & 4.28 & 3.04\\
        \hline
         \multicolumn{2}{c|}{VIS-LWIR} & 8.71 & 7.19	& 4.42 & 3.07 & 4.38 & 5.41	& 3.3 & 8.49 & 5.63 \\
         \multicolumn{2}{c|}{Optical-SAR} &  17.52 & 15.81 & 17.05 & 16.97 & 15.68 & 17.86 & 14.89 & 21.81 & 17.20 \\
         \multicolumn{2}{c|}{VL-CMIM} & 8.94  & 6.69 &  2.30 & 2.78 & 2.25 & 4.09 & 2.33 & 7.34 & 4.59 \\
         
        \hline
    \end{tabular}}}
     \label{tab:scca-VIS-NIR-transfer-learning}
\end{table*}

\begin{table*}[ht]
\renewcommand\arraystretch{1.3}
    \centering
    \caption{Cross-dataset Transfering Performance: trained on other datasets and test on the VL-CMIM dataset}
    \setlength{\tabcolsep}{1.5mm}{
    \scalebox{0.98}{
    \begin{tabular}{ l l |  c c c c c  |c }
    \hline
         \multicolumn{2}{c|}{\multirow{2}*{\diagbox{Train Dataset}{Test Dataset}}} &  \multicolumn{5}{c|}{VL-CMIM} & \multirow{2}*{Mean}\\
        \cline{3-7}
        &  &  Asteroid & Field & Build & Street & Water  \\
        \hline
        \multirow{3}*{Brown} & Yosemite & 4.87 & 18.12 & 9.54 & 14.41 & 12.14  & 11.82 \\
        & Notredame & 4.29 & 18.43 & 9.98 & 14.11 & 12.94 & 11.95  \\
        & Liberty & 5.11  & 16.39 & 9.52 & 12.02 & 11.30 & 10.87 \\
        \hline
         \multicolumn{2}{c|}{VIS-LWIR} & 8.73 & 13.59 & 8.72 & 14.73 &  9.47 & 11.05 \\
         \multicolumn{2}{c|}{Optical-SAR} & 12.57 & 10.14 & 9.81 & 10.56 & 9.64 & 10.54 \\
         \multicolumn{2}{c|}{VIS-NIR} & 3.74 & 7.70 & 4.87 & 6.65 & 5.47 & 5.69   \\
         
        \hline
     \end{tabular}}}
    \label{tab:vl-cmim transfer learning}
\end{table*}

\begin{table}[ht]
\renewcommand\arraystretch{1.3}
    \centering
    \caption{Cross-dataset Transfering Performance: trained on other datasets and test on VIS-LWIR and Optical-SAR dataset}
     \scalebox{0.9}{
    \begin{tabular}{c c| c c}
    \hline
    \multicolumn{2}{c|}{\diagbox{Train Dataset}{Test Dataset}} & VIS-LWIR & Optical-SAR\\
    \hline
    \multirow{3}*{Brown} & Yosemite & 10.99 & 13.53\\
                        & Notredame & 10.74 & 16.12\\
                        & Liberty & 13.17 & 18.49 \\    
    \hline
    \multicolumn{2}{c|}{VIS-NIR} & 6.10 & 9.76\\
    \multicolumn{2}{c|}{VIS-LWIR} & - & 25.19\\
    \multicolumn{2}{c|}{Optical-SAR} & 23.76 & - \\
    \multicolumn{2}{c|}{VL-CMIM} & 7.60 & 25.57\\
    \hline
    \end{tabular}}
    \label{tab:icip and Optical-SAR transfer learning}
\end{table}

\subsection{Comparison with the State-of-the-Art}

1) Results on VIS-NIR dataset: In order to demonstrate the effectiveness and generalization of our network on multi-modal image patch matching tasks, we compare the proposed method with the state-of-the-arts image matching methods on VIS-NIR dataset. The result as shown in Table \ref{tab:scca-compare-vis-nir}.

Our method outperforms the state-of-the-art AFD-NET \cite{quan2019afd} and other comparison methods. The main reason is that our network focuses on deep interaction of spatial-correlation features and channel-wise attention features and thus can reduce the differences between multimodal images and extract cross-domain invariant features. Among these works, SIFT\cite{lowe2004distinctive}, GISIFT\cite{firmenichy2011multispectral}, EHD\cite{aguilera2012multispectral}, LGHD \cite{shechtman2007matching} are all traditional hand-designed descriptor methods, which are not suitable for large differences multimodal images. PN-NET \cite{balntas2016pn}, Q-Net \cite{savinov2017quad}, L2-Net \cite{tian2017l2} and HardNet \cite{mishchuk2017working} are all descriptor learning methods. They focus on learning a representation that can enable the two matched features as close as possible, while non-matched features are far apart. The PN-NET is the first applied to cross-spectral image patch matching task. Siamese \cite{simo2015discriminative}, Pseudo-Siamese \cite{zagoruyko2015learning}, 2-Channel \cite{zagoruyko2015learning}, SCFDM \cite{quan2019cross}, Hybrid \cite{baruch2021joint} Moreshet \& K+ \cite{moreshet2021paying}, Quan \& W+ \cite{quan2021multi}, AFD-Net \cite{quan2019afd} and MFD-Net \cite{yu2022multi} are all metric learning methods. They focus on designing feature extraction networks and achieved significant performances. 


2) Results on Brown Dataset: Brown image dataset is a single model dataset. as shown in Table \ref{tab:scca-compare-brown-1}, Our method outperforms other comparison methods. Compared with the current state-of-the-art method, the mean FPR95 is significantly improved by 0.63\%. Experimental results demonstrate that our method is not only suitable for cross-modality datasets, but also performs well in single-modality image matching tasks.

3) Results on Optical-SAR Dataset and VIS-LWIR Dataset:
On the two datasets, we compared the method in this chapter with five existing image patch matching methods, and the experimental results are shown in Table \ref{tab:scca-compare-vis-lwir}.
Optical-sar includes optical images and synthetic aperture radar images. 
There are significant differences between them in appearance, and it is difficult to match them. Our method has achieved good performance on this dataset (FPR95=0.78), which is far superior to other methods. The significant performance improvement shows that DPI-QNet has good generalization performance in multi-modal image matching tasks. On VIS-LWIR dataset, the method in this chapter has significantly improved compared with other methods in FPR95, a major index. Because the DPI-QNet network proposed in this paper can learn the features related and unrelated to the imaging mechanism between multi-modal images, the network can extract more discriminative features, and the average FPR95 is 1.97.


4) Results on VL-CMIM Dataset:
VL-CMIM is the first large visible and long-wave infrared spectrum image patch dataset. Similar to VIS-NIR dataset, we trained on the country category and tested on other categories, which can well verify the generalization performance of the network. The experimental results are shown in Table \ref{tab:scca-compare-vis-vl-cimi}. The average value of FPR95 on the VL-CMIM dataset of DPI-QNet is 3.95, and the performance is far superior to other methods.

\subsection{Ablation study}


To verify the effectiveness of each module in the multi-modal image patch matching network (DPI-QNet) based on spatial correlation and channel-wise attention with deep interaction, we conduct ablation experiments on the VL-CMLM dataset. The experimental results were shown in Table \ref{tab:ablation-scca}, where "Sia" and "Pre-Sia" means siamese network and pseudo-siamese network, respectively. "CL" is spatial correlation. "EPSA" means the pyramid splits attention module. "TF1" and "TF2" respectively represent transformer module based on spatial correlation feature extraction module and multi-scale channel attention feature aggregation module. "DI" means deep interactive fusion. $\mathbf{SCCA}^\mathrm{T}$ means "CL", "EPSA" and "TF1," "TF2" change order.

The experimental results show that the mean value of FPR95 decreases by 1.59 when the spatial correlation module is added to the baseline network separately. When the pyramid split attention module was added to the baseline network alone, the mean value of FPR95 was reduced by 0.02. When spatial correlation module and pyramid splitting attention module were added to the baseline network, the average value of FPR95 decreased by 2.38, and the experimental effect was significantly improved.


Secondly, to prove Transformer effectiveness in spatial correlation and pyramid splitting attention, we only consider spatial correlation and pyramid splitting attention or Transformer encoder. The experimental results show that the mean value of FPR95 decreases by 1.22 when spatial correlation and pyramid splitting attention are added to the benchmark network alone, compared with the benchmark network. When only Transformer encoder architecture is used in the measurement fusion phase, the results are not good, with an average FPR95 of 8.94. However, when Transformer, spatial correlation and pyramid splitting attention are added to the benchmark network at the same time, the average value of FPR95 decreases by 2.38, and the experimental effect is significantly improved.


Transformer can expand the receptive field of the image and obtain the global context of the image. 
To prove the effectiveness of Transformer in spatial correlation and pyramid splitting attention, we consider comparing two sequences, the sequence one is "CL", "EPSA", "TF1", "TF2"  and the sequence two is "TF1", "TF2", "CL", "EPSA". The experimental results show that compared with the benchmark network, the FPR95 of sequence one is improved by 2.38, and that of sequence two is improved by 0.94.


The fourth part is the deep interaction fusion module, which aims to deeply interact the extracted information from the two dimensions of space and channel, promote the flow of information between multiple modes, and enable the network to better learn the cross-domain consistency characteristics.
For the appearance differences in cross-modal images, this chapter also explores the deep interactive fusion strategy. Through the fusion of deeper numbers, the information flow between modes is promoted, so that each branch can not only pay attention to its own discriminant features, but also learn the consistency features between other modal images, so as to improve the accuracy of cross-modal image patch matching.


Finally, the deep interactive fusion module can promote the flow of information between modes through the fusion of deeper numbers, so that each branch can not only pay attention to its own discriminant features, but also learn the consistency features between images of another mode. In order to prove the effectiveness of this module, we add the deep interactive fusion module on the basis of sequence one. The average value of FPR95 decreased by 0.36, indicating significant experimental performance.

\subsection{Cross-dataset Transferring Performance}



To evaluate the cross-dataset transferring performance of the proposed method, transferring learning was validated in four existing publicly available image patch matching datassets, including VIS-NIR(Visible and near infrared), VIS-LWIR(visible and Long-wave Infrared Image), Optical SAR(visible and synthetic aperture radar), and Brown (single-modal) dataset. Specific instructions are as follows:

Firstly, the training models of VIS-LWIR, Optical-SAR and Brown datasets were tested on VIS-NIR dataset. The experimental results are shown in Table \ref{tab:scca-VIS-NIR-transfer-learning}. The three subsets in Brown, VIS-LWIR and Optical-SAR datasets achieved an average FPR95 of 2.61, 2.57, 3.04, 5.63 and 17.20, respectively, in VIS-NIR datasets.
As the imaging mechanism and characteristics of SAR images are quite different from those of other modal images, the generalization effect is poor, which is the same as our initial assumption. Except for the models trained on the Optical-SAR dataset, other models achieved relatively good performance.



Secondly, VIS-LWIR and Optical-SAR datasets were selected as testsets, and the training models of the other three datasets were used for testing on VIS-LWIR and Optical-SAR data sets respectively. The experimental results are shown in Table \ref{tab:tab:icip and Optical-SAR transfer learning}. The model trained in VIS-NIR dataset performed better on VIS-LWIR dataset and Optical-SAR dataset, with the average FPR95 reaching 6.10 and 9.76, respectively.

Finally, the Brown dataset is used as the test. The experimental results are shown in Table \ref{tab:Brown-transfer-learning}. The average FPR95 of VIS-NIR, VIS-LWIR and Optical-SAR data sets in Brown dataset reached 2.15, 2.86 and 3.68, respectively. Experiments show that the network has good generalization performance and robustness.

\section{CONCLUSION} \label{conclusion}

In this paper, firstly, we construct a first large VIS-LWIR image patch dataset VIS-LWIR suitable for multimodal image patch matching, which strictly aligned in time and space. Secondly, we proposed a novel network DPI-QNet for multimodal image matching, which could efficiently interact of spatial-correlation features and channel-wise attention features. It can well reduce the differences between multimodal images and extract cross-domain invariant features. The experiments show that our DPI-QNet outperformed other methods on the VL-CMIM, VIS-LWIR, VIS-NIR, Optical-SAR and Brown dataset.




 \bibliographystyle{elsarticle-num} 
 \bibliography{cas-refs}





\end{document}